\documentclass[lettersize,journal]{IEEEtran}
\usepackage{amsmath,amsfonts}
\usepackage{algorithm}
\usepackage{array}
\usepackage{textcomp}
\usepackage{stfloats}
\usepackage{url}
\usepackage{verbatim}
\usepackage{graphicx}
\usepackage{cite}
\usepackage{amssymb}
\usepackage{booktabs}
\usepackage[table]{xcolor}
\usepackage[utf8]{inputenc}
\usepackage[noend]{algpseudocode}
\usepackage{multirow}
\usepackage{subfigure}
\usepackage{orcidlink}
\hypersetup{hidelinks, colorlinks, linkcolor=blue, citecolor=blue, urlcolor=blue}
\usepackage{picinpar}   
\usepackage{diagbox}

\def\BibTeX{{\rm B\kern-.05em{\sc i\kern-.025em b}\kern-.08em
    T\kern-.1667em\lower.7ex\hbox{E}\kern-.125emX}}
\usepackage{balance}
\usepackage{threeparttable}

\begin{document}

\title{RoCA: Robust Contrastive One-class Time Series Anomaly Detection with Contaminated Data}

\author{Xudong Mou$^{\orcidlink{0009-0005-1445-3742}}$, Rui Wang$^{\orcidlink{orcid=0009-0004-5900-605X}}$, Bo Li$^{\orcidlink{0000-0002-1577-0310}}$,~\IEEEmembership{Member, IEEE}, Tianyu Wo$^{\orcidlink{0000-0002-5331-3364}}$,~\IEEEmembership{Member, IEEE}, \\ Jie Sun$^{\orcidlink{0000-0002-5952-1850}}$, Hui Wang, and Xudong Liu$^{\orcidlink{0000-0001-8566-660X}}$,~\IEEEmembership{Member, IEEE}
\thanks{ (Bo Li and Tianyo Wo are the corresponding authors)}
\thanks{
Xudong Mou, Rui Wang, Bo Li, Hui Wang, and Xudong Liu are with School of Computer Science and Engineering, Beihang University, Beijing, China. (Email: mxd@buaa.edu.cn, ruiking@buaa.edu.cn, libo@act.buaa.edu.cn, whui@buaa.edu.cn, liuxd@buaa.edu.cn)
}
\thanks{Jie Sun and Xudong Liu are with Zhongguancun Laboratory, Beijing, China. (Email: sunjie@zgclab.edu.cn) }
\thanks{
Tianyu Wo is with School of Software, Beihang University, Beijing, China. (Email: woty@buaa.edu.cn)
}
\thanks{Manuscript received March 2, 2025.}}

\markboth{Journal of \LaTeX\ Class Files,~Vol.~14, No.~8, August~2021}%
{Shell \MakeLowercase{\textit{et al.}}: A Sample Article Using IEEEtran.cls for IEEE Journals}

\IEEEpubid{0000--0000/00\$00.00~\copyright~2021 IEEE}

\maketitle

\begin{abstract}
The accumulation of time-series signals and the absence of labels make time-series Anomaly Detection (AD) a self-supervised task of deep learning. Methods based on normality assumptions face the following three limitations: (1) A single assumption could hardly characterize the whole normality or lead to some deviation. (2) Some assumptions may go against the principle of AD. (3) their basic assumption is that the training data is uncontaminated (free of anomalies), which is unrealistic in practice, leading to a decline in robustness. 
This paper proposes a novel robust approach, RoCA, which is the first to address all of the above three challenges, as far as we are aware. It fuses the separated assumptions of one-class classification and contrastive learning in a single training process to characterize a more complete so-called normality. Additionally, it monitors the training data and computes a carefully designed anomaly score throughout the training process. This score helps identify latent anomalies, which are then used to define the classification boundary, inspired by the concept of outlier exposure. The performance on AIOps datasets improved by 6\% compared to when contamination was not considered (COCA). On two large and high-dimensional multivariate datasets, the performance increased by 5\% to 10\%. RoCA achieves the highest average performance on both univariate and multivariate datasets.
The source code is available at \url{https://github.com/ruiking04/RoCA}.

\end{abstract}


\section{Introduction}
\label{introduction}
\IEEEPARstart{T}{ime} Series Anomaly Detection (TSAD) refers to identifying abnormal samples that significantly deviate from most time series data~\cite{grubbs1969procedures}. It increasingly contributes to a variety of applications like healthcare, manufacturing, and intrusion detection to prevent potential risks. With the development of sensors, the Internet of Things (IoT), and cloud/edge computing, the accumulation of time series data can be easier than before, which engages the performance of deep learning superior to shallow ones ~\cite{pang2021deep}. Since labeling the outlier from quantities of temporal data could be costly and tricky, TSAD is usually considered an unsupervised learning problem in which learning representation for discerning anomalies relies on some normality assumptions. They assume that the so-called ``normality" should possess certain attributes, which can be described with autoencoder~\cite{malhotra2016lstm,mou2023deep}, one-class classification (OC)~\cite{xu2024calibrated,chong2020simple,ruff2018deep}, contrastive learning (CL)~\cite{de2021contrastive,sohn2020learning}, and so on. Taking autoencoder-based methods as an example, they reconstruct the original time series and calculate the reconstruction loss, assuming that normal samples are better restructured (get lower losses) from the latent space than abnormal ones. However, these methods still suffer from three serious limitations. \IEEEpubidadjcol

\begin{figure*}[ht!]
  \centering
  \includegraphics[width=0.85\linewidth]{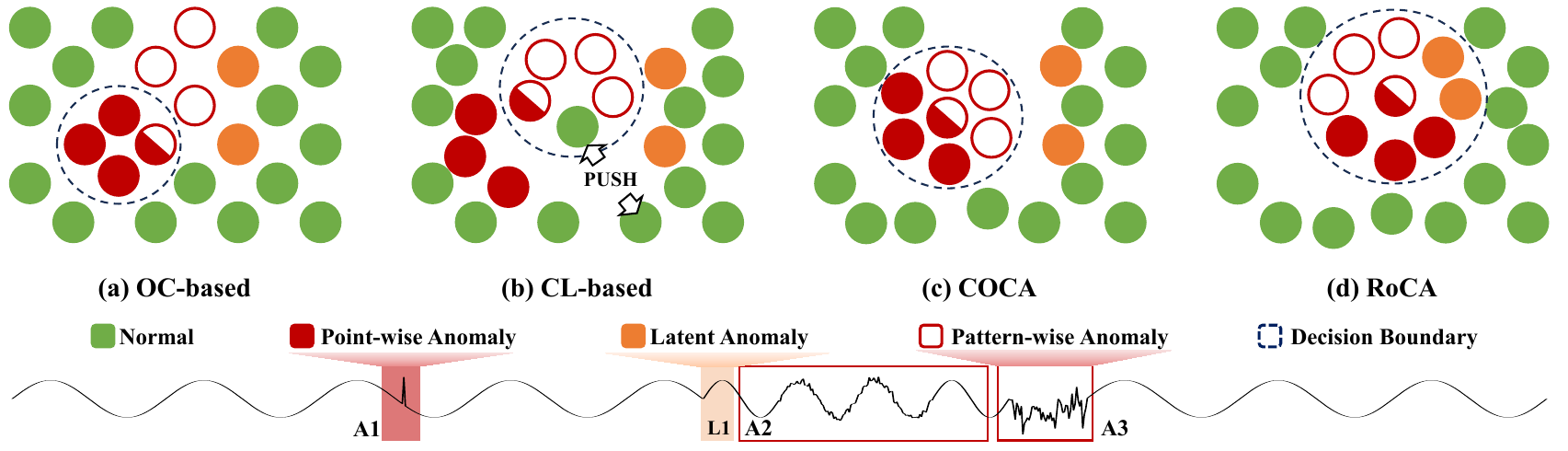}
  \caption{Motivation Diagram. In time series data, various types of anomalies exist, including point-wise anomalies (e.g., A1) and pattern-wise anomalies (e.g., A2, A3). Moreover, due to untimely detection in the early stages or data contamination, some signs of anomalies may not be identified and labeled. For instance, the segment preceding A2 may exhibit such characteristics. We refer to these as latent anomalies (L1). Methods based on single assumptions can only detect specific anomalies (a, b). COCA combines multiple assumptions (c) to enhance the model's anomaly detection capability. RoCA (d) identifies and leverages the contaminated data or latent anomalies to optimize the detection boundary.
}
  \label{motivation}
\end{figure*}

\begin{enumerate}

\item[\textbf{L1:}]  A single normality assumption may be one-sided, often derived from the pretext task of self-supervised representation learning. In addition, time-series anomalies come in various forms, including point-wise anomalies (global or local) and pattern-wise anomalies of shapelet, trend, and season~\cite{lai2021revisiting}. Therefore, it is insufficient to detect all of them based solely on one normality assumption, as shown in Fig.\ref{motivation} (a) and (b). 
\item[\textbf{L2:}] Some assumptions may go against the very nature of the anomaly detection task, that is, extracting features common to the vast majority of normal samples. For instance, CL-based ones~\cite{de2021contrastive, oord2018representation, qiu2021neural} assume that there is more mutual information between normal contrastive pairs than anomalous ones. Comparison pairs are ``positive" (views of the same sample) or ``negative" (views of different samples). Positive pairs are pulled together, while negative pairs are pushed apart. However, if both samples in a negative pair are normal, they would be misseparated, failing to capture shared class information, as in \cite{caron2020unsupervised}. Therefore, the negative pairs in CL-based AD methods may contradict the core principle of anomaly detection, resulting in a false positive, as illustrated in Fig. \ref{motivation} (b).
\item[\textbf{L3:}] These normality-based approaches are typically grounded in the assumption that the training data are uncontaminated, containing little to no anomalies. This fundamental assumption is unrealistic in most applications and poses a significant challenge to the robustness of AD models. In many cases, large, uncurated datasets may already contain the latent anomalies being targeted, such as the undetected early stages of network intrusions. As the proportion of latent anomalies increases, the effectiveness and robustness of AD models decrease. As shown in Fig.\ref{motivation}, most methods (a,b, and c) have difficulty identifying these potential anomalies. Previous work~\cite{ruff2018deep, beggel2020robust, yoon2021self, xu2022calibrated} training with polluted data often blurred the classification boundary or wasted the knowledge contained in hidden anomalies, and were only verified with single assumptions. 
\end{enumerate}

To overcome the above limitations, this paper proposes RoCA, a novel robust time series anomaly detection method based on multi-assumptions.  It seamlessly combines the assumptions of CL and one-class classification into a binding one with a well-designed loss term, rather than regarding it as a multi-task learning issue. It treats the original and reconstructed representations as a positive pair of negative-sample-free CL, referred to as ``sequence contrast". During training, an outlier exposure loss term is applied to characterize anomalies, dynamically identifying potential abnormal or contaminated training samples and pushing them away to refine the classification boundary.
Our prior work, COCA~\cite{wang2023deep} (Fig.~\ref{motivation} c), overcomes the limitations of a single assumption and explores finding anomalies in univariate time series with fused assumptions. It lacks robustness due to not addressing the essential issue of methods based on normal assumptions. On the other hand, it has not been validated on multivariate datasets. In this paper, we extend COCA in several key ways:

\begin{itemize}
\item  We propose the robust RoCA, the first attempt at a multi-assumption based anomaly detection method for contaminated data, to our best knowledge. It integrates CL and one-class classification, provides a more comprehensive characterization of normality, and defines a clearer detection boundary by leveraging the often-overlooked but valuable latent anomalies within the training data.
\item We propose a new negative pair free time-series CL paradigm, namely ``sequence contrast". We clarify that the essence of contrastive learning lies in distinguishing representations, rather than the compared pairs or the negative samples.
\item We conduct comprehensive experiments and ablation studies that offer
deeper insights into the model’s effectiveness and robustness across uni- and multi-variate time series datasets with different degrees of contamination. Performance on AIOps datasets improved by 6\% compared to when contamination was not considered. For multivariate time series data that are prone to mislabeling and noise, performance increased by 5\% to 10\%.
\end{itemize}

The remainder of this paper is organized as follows. 
First, we summarize the related work about contrastive learning, time series anomaly detection based on deep learning, and explorations in training with contaminated data in Section~\ref{related}. Section~\ref{Preliminary} presents the problem definition, including the key notations and their meanings, along with the fundamental principles of one-class classification and contrastive learning utilized in this study. Subsequently, in Section~\ref{Sec3}, we provide a detailed explanation of the proposed methodology, covering the overall framework and our carefully designed loss function, which comprises three essential components. 
Then, Section~\ref{Sec4} reports and analyzes the experimental results on various datasets with different levels of contamination. Finally, we conclude this paper and discuss future work in Section~\ref{section conclusion}.

\section{Related work}
\label{related}
Our research involves contrastive learning, deep anomaly detection, and training along with contaminated data. We will review recent related work from the three aspects.
\subsection{Contrastive learning}
The recent resurgence of academic attention in contrastive learning began with CPC~\cite{oord2018representation}, which pulls positive samples closer and distances negative samples with the InfoNCE loss. It obtained great performance in representation learning, though relying on a large number of negative samples. Then \cite{wang2020understanding} clarifies two key properties of CL: (1) alignment, i.e., similar samples should carry similar representations (pull positive pair), and (2) uniformity: representations follow a uniform distribution on a hypersphere (push negative pair). In the following research, BYOL~\cite{grill2020bootstrap}, SwAV~\cite{caron2020unsupervised}, and SimSiam~\cite{chen2021exploring} achieved the property of uniformity without using negative samples, which inspired us to design our loss function.
Meanwhile, SimCLR~\cite{chen2020simple} and TS-TCC~\cite{eldele2021time} learn invariant representations for visual data and time series by aligning augmented data representations. In addition, TS-TCC uses a temporal contrasting module to address the temporal dependencies of time series, showing us the possibility of different contrastion objects.
Although all these CL approaches have successfully improved representation learning in the domain of visual data and time series, they could be inapplicable to time-series AD. For example, contradictions exist between the uniformity of contrastive learning and the nature of class imbalance in AD.

\subsection{Deep anomaly detection}
Deep anomaly detection methods can generally be categorized into two kinds: deep learning for feature extraction and learning feature representations of normality~\cite{pang2021deep}.
The former is a two-staged learning method that uses deep approaches to obtain representations and passes them to downstream anomaly detection.
However, it does not address the downstream task directly, thus the pre-trained representations may be detrimental to anomaly detection.
The later group couples representations learning with anomaly scoring in some way, such as GANs-based~\cite{schlegl2017unsupervised}, autoencoder-based~\cite{malhotra2016lstm}, one-class classification-based~\cite{ruff2018deep}, clustering-based~\cite{zong2018deep}, saliency map-based~\cite{ren2019time}, multi-transformation based~\cite{han2024self}, diffusion based~\cite{basak2024diffusion} and contrastive learning-based~\cite{qiu2021neural,de2021contrastive} methods.
The effectiveness of these methods hinges on their underlying assumptions about normality and anomaly, some of which are inspired by the pretext tasks in self-supervised learning.
For instance, GANs-based methods operate on the assumption that normal samples can be more accurately generated from the latent space of the generative network compared to anomalies.
However, these assumptions about normal samples may only address a single dimension of overall normality.
Uniquely, RoCA stands out by avoiding pre-training and seamlessly integrating the normality assumptions from one-class classification and contrastive learning to identify anomalies in time-series data. 

\begin{table*}[!ht]
\caption{Notations and Descriptions.}
\label{notations}
\renewcommand{\arraystretch}{1}
\centering
\scalebox{1}{
\begin{tabular}{cl  cl}
\toprule
 Notation  & Description & Notation  &  Description\\ 
\midrule
 $\mathcal{S}$  & An ordered time series. &  $x_t$ & Vector collected at timestamp $t$. \\
 $T$  & Time period of whole $\mathcal{S}$. &  $dim$ & Dimension of $x_t$. \\
 $\mathcal{D}$  & Dataset of $\mathcal{S}$ divided by a sliding window. &  $L$ & Length of sliding window. \\
$\delta$  & Time step of sliding. &  $X_i$ & The $i^{th}$ sample in $\mathcal{D}$. \\
$y_i$  & Label of $X_i$.  & $\mathcal{Y}$  & Set of labels.  \\
$f_\Theta$  & Encoder  &   $g_\Theta$  & Seq2Seq Encoder. \\
$h_\Theta$  & Decoder. &  $p_\Theta$  & Projector. \\
$z$  & Representation of original sample. &  $z^{\prime}$  & Representation of reconstruction. \\
$q$  & Projection. &  $q^{\prime}$  & Projection of reconstruction.  \\
$c_L$  & Contextual representation. &  $Ce$  & One-class center. \\
$\alpha$  & One-class error of $q$ on hypersphere. &  $\beta$  & One-class error of $q^{\prime}$ on hypersphere. \\
$\gamma$  & Constrastive error on hypersphere. &  $\theta$  & The dihedral angle between $CeOq$ and $CeOq^{\prime}$.\\
$\mathcal{L}_{svdd}$  & Loss of SVDD. &  $\mathcal{L}_{sim}$  & Loss of Negative-sample-free contrastive learning. \\
$\mathcal{L}_{RoCA}$  & Loss of RoCA. & 
$\mathcal{L}_{COCA}$  & Loss of COCA.  \\
$\mathcal{L}_{COCAS}$  & Loss of RoCA. & 
$\mathcal{L}_{Inv}$  & Invariance term. \\
$\mathcal{L}_{OE}$  & Outlier exposure term. &  $\mathcal{L}_{Var}$  & Variance term. \\
$\mathcal{L}_{joint}$  & Joint loss of $\mathcal{L}_{Inv}$ and $\mathcal{L}_{OE}$. &     $\varepsilon$  & A small scalar to prevent instabilities.\\
$\nu$  & Proportion of contaminated data. &  $\mu$  & Weight of $\mathcal{L}_{OE}$. \\
$\lambda$  & Weight of $\mathcal{L}_{Var}$. &  $\zeta$  & Constant target value of the standard deviation. \\
$S_i^{train}$ & Anomaly score of $X_i$ during training. &  $S_i^{test}$ & Anomaly score of $X_i$ during testing. \\
\bottomrule
\end{tabular}}

\end{table*}

\subsection{Training with contaminated data}
Currently, the approaches of training along with contaminated data fall into three categories, including ``soft-boundary", ``refine", and ``LOE". 
\emph{Soft-boundary} refers to adjusting the hinge losses to allow a portion of samples to cross the classification border~\cite{wang2023deep, ruff2018deep}, indicating that those are anomalies and do not contribute to normality. However, the boundary is too lenient to separate the normal samples from the abnormal ones. \emph{``Refine"}~\cite{beggel2020robust,  yoon2021self, xu2022calibrated} takes a pre-train process to select suspicious samples according to the losses and sets their weights to 0 while retraining. \cite{lin2024exploiting} uses a noise term to weaken the impact of suspicious samples on the model optimization. However, the rare knowledge about the anomalies is wasted.
To leverage the potential of unknown anomalies, \emph{LOE}~\cite{qiu2022latent} draws inspiration from outlier exposure~\cite{hendrycks2018deep} and extracts learning from both normal and abnormal data, thereby achieving notable improvements. However, its effectiveness has yet to be verified on time series and pure datasets, nor has it been tested with multi-assumption approaches. This work has nevertheless inspired us to incorporate an outlier exposure term into the loss function to identify and use the latent anomalies.

\section{Preliminary}
\label{Preliminary}
In this section, we present the problem of TSAD and the task of training with contaminated data, the method of one-class classification, and contrastive learning without negative samples.

\subsection{Problem definition}
Given an ordered time series $\mathcal{S}=\left\{x_1, x_2,\dots,x_T\right\}$ collected during time $T$, and $x_{t}\in \mathop{\mathbb{R}}^{dim}$ is a $dim$-dimensional vector collected at timestamp $t$. 
$\mathcal{S}$ is univariate or multivariate when $dim=1$ or $dim>1$. 
As a routine approach, $\mathcal{S}$ is split into an unlabeled time subsequence set $\mathcal{D}=\left\{{\bf X}_1,{\bf X}_2,\dots,{\bf X}_N\right\}$ by length-$L$ sliding windows with time-step $\delta \leq L$, where ${\bf X}_i=\left\{x_1,x_2,\dots,x_L\right\}$ is a subsequence sample and $N$ is the number of samples. 
The length-$L$ sliding window is overlapping if $\delta<L$. 
Each sample ${\bf X}_i$ has an unknown binary label $y_i\in \{0,1\}$, indicating that $\bf{X}_i$ is normal (0) or anomalous (1).  In time-series AD, the anomaly score $\mathcal{S}_{i}$ of ${\bf X}_i$ is calculated by the AD model such that the higher $\mathcal{S}_{i}$ is, the more likely it is an anomalous time series. 

Suppose $\nu$ proportion of samples in $\mathcal{D}$ are abnormal.
This paper aims to train normality assumption-based anomaly detectors on such corrupted data. The challenge is to infer the binary labels $y_i$ during training while exposing the $\nu$ of anomalies to the model for a clear boundary. Denote the set of labels within $\mathcal{Y}$.
For clarity and reference, we elaborate on the notations and definitions used in this paper, as shown in Table~\ref{notations}.

\begin{figure*}[ht]
  \centering
  \includegraphics[width=\linewidth]{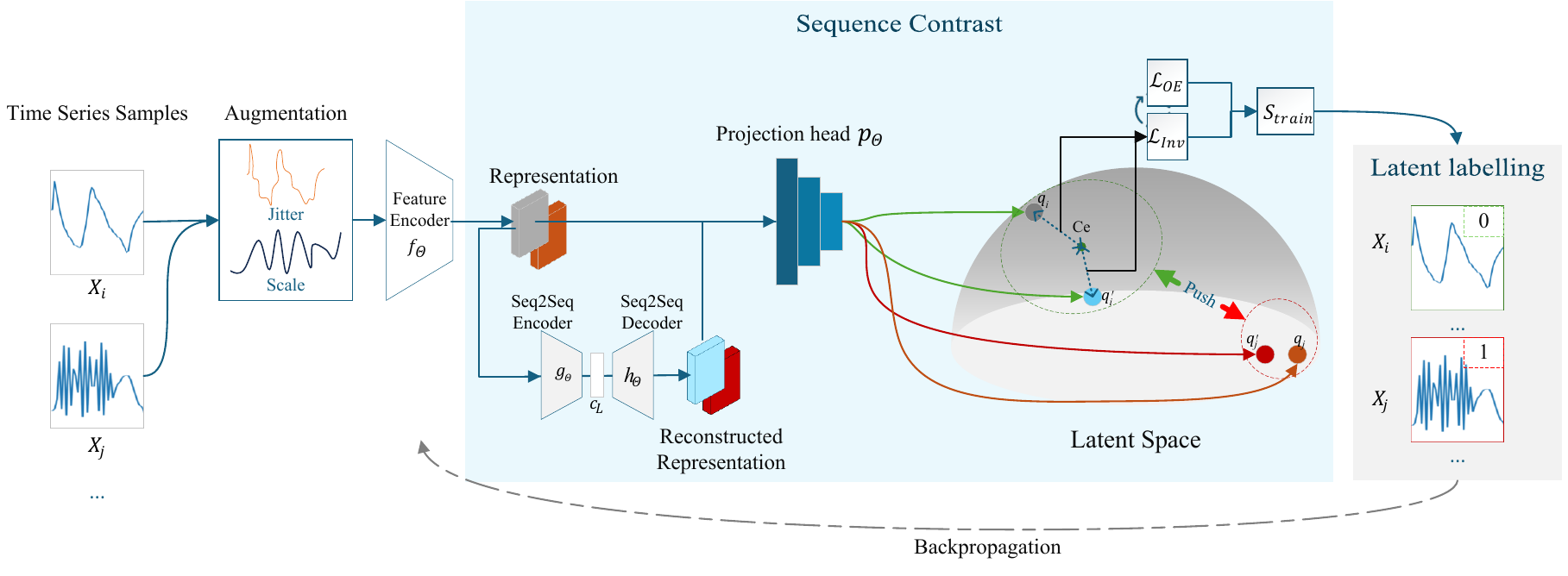}
  \caption{The framework of the RoCA. }
  \label{architecture}
\end{figure*}

\subsection{One-class classification.}
The optimization objective of Deep SVDD \cite{ruff2018deep}, a representative method for one-class classification, is defined as:
\begin{equation}
\mathcal{L}_{svdd} =\frac{1}{N}\sum_{i=1}^{N}\|\phi(x_{i},\Theta)-Ce\|^{2},
\label{Deep_SVDD_loss}
\end{equation}
where $Ce \in \mathcal{Z}$ is the one-class center, $\Theta$ is the set of parameters of a representation network $\phi$.
Deep SVDD obtains the sphere of the smallest volume by minimizing the $\mathcal{L}_{svdd}$ in the representation space $\mathcal{Z} \subset \mathop{\mathbb{R}}^{K}$.

\subsection{Negative-sample-free contrastive learning.}
BYOL \cite{grill2020bootstrap}, SimSiam \cite{chen2021exploring}, and Vicreg \cite{bardes2022vicreg} are representatives of contrastive learning without negative pairs.

The optimization objective of SimSiam is simplified as:
\begin{equation}
\mathcal{L}_{sim} =\frac{1}{N}\sum_{i=1}^{N}-\frac{z_{i}}{\Vert z_{i} \Vert_{2}} \cdot \frac{z_{i}^{\prime}}{\Vert z_{i}^{\prime} \Vert_{2}},
\label{cl_loss_2}
\end{equation}
where $z_{i}$ and $z_{i}^{\prime}$ are the representations of contrasting objects (positive pairs) in the latent space $\mathcal{Z}$.
Equation (\ref{cl_loss_2}) is essentially pulling the positive pair close using cosine similarity.
As for the ``hypersphere collapse" caused by no negative pairs, BYOL and SimSiam solve it by bootstrap and asymmetric networks, and Vicreg by variance.

\section{Methodology} \label{Sec3}
The key insight of RoCA lies in merging multiple normality assumptions into a binding one, and the samples that violate it may be latent anomalies. To address the three key limitations mentioned in the introduction, RoCA involves three key elements: fused assumption of CL and one-class classification, negative-free sequence contrast, and an iterative training process based on temporary anomaly scores during training.
In this section, we propose the overview of RoCA in Part A, then explain the vital loss function in Part B, and introduce the inference process in Part C.

\subsection{RoCA framework overview}


As shown in  Fig.~\ref{architecture}, time series ${\bf X}_i$ is one of an augmented sample of the raw dataset. A multi-layer temporal convolution feature encoder $f_{\Theta}:\mathcal{X}\mapsto\mathcal{Z}$ takes ${\bf X}_i$  of length $T$ as the input and outputs the latent representations $z_{1},\dots, z_{L}$ for $L$ time-steps, potentially with a lower temporal resolution, i.e. $T>L$.
Next, the representations are fed to a Seq2Seq encoder $g_{\Theta}:\mathcal{Z}\mapsto\mathcal{C}$ to summarize all $z_{\leq L}$ as context vectors $c_{L}$.
Then a Seq2Seq decoder $h_{\Theta}:\mathcal{C}\mapsto\mathcal{Z}$ outputs reconstructions $z^{\prime}_{1},\dots, z^{\prime}_{L}$ for $L$ time-steps to learn temporal dependencies.
Furthermore, original representations together with the reconstructed ones are fed to a learnable nonlinear projector $p_{\Theta}:\mathcal{Z}\mapsto\mathcal{Q}$ to obtain the projections $q$ and $q^{\prime}$.
The projections are then passed to calculate the loss (see next Subsection~\ref{Objective}) to maximize the similarity between $q$ and $q^{\prime}$ concerning the one-class center $Ce \in \mathcal{Q}$, combining the two normality assumptions of contrastive learning and one-class classification-based. During the process, we adopt an overlapping update of the latent anomaly labels and parameters to push possible anomalies away, clarifying the boundary of normal and contaminated data.

\textbf{Time-Series Augmentation.}
Data augmentation enhances the performance of anomaly detection (AD) methods by increasing the training data volume and making it easier to isolate anomalies \cite{sohn2020learning}. In this paper, we apply jittering (noise addition) and scaling (pattern-wise magnitude change) to expand the training set. It is crucial to select the jittering and scaling hyper-parameters carefully, considering the specific characteristics of the time-series anomalies.

\textbf{Feature encoder.}
The encoder network accepts normalized time series input with zero mean and unit variance.
It has a multi-block temporal convolutional architecture, 2 blocks for AIOps and UCR, 3 for SWaT and WADI, since univariate data has simpler features than multivariate data, we use an extra layer of blocks for the latter to capture more representation information. Each block comprises a Conv1D layer, a BatchNormalization (BN) layer, a ReLU activation function, and a MaxPool1D layer, where the first block also contains a Dropout layer. 

\textbf{Seq2Seq.}
The Seq2Seq component comprises an encoder and a decoder, each implemented as a 3-layer Long Short-Term Memory (LSTM) network. The decoder is followed by a fully connected (FC) layer. In this paper, the length of the hidden space representation is less than 20, making LSTM suitable for capturing the context representation. However, recent advancements in Seq2Seq modeling for longer sequences, such as self-attention networks or the Transformer model, could further enhance the results.

\textbf{Projector.}
The projector employs an MLP with one hidden layer with BN and ReLU activation to map representations into the space where contrastive one-class loss is computed.

\subsection{The optimization objective \label{Objective}}
The RoCA optimization objective consists of \textit{invariance}, \textit{outlier exposure}, and \textit{variance} terms.
The invariance term $\mathcal{L}_{Inv}$ is to maximize the cosine similarity between the one-class center $Ce$, representations $q_{i}$, and seq2seq outputs $q_{i}^{\prime}$ in the projection space $\mathcal{Q}$. The outlier exposure term $\mathcal{L}_{OE}$ pushes the latent outliers far away from $Ce$. Then the variance term $\mathcal{L}_{Var}$ avoids ``hypersphere collapse'' without negative sample pairs.

\subsubsection{Invariance term}
A crude way to integrate one-class classification and contrastive learning is treating it as multi-task learning with adjustable hyper-parameters as follows:
\begin{equation}
param_1 \cdot \mathcal{L}_{svdd} + param_2 \cdot \mathcal{L}_{sim}.
\label{multi_task}
\end{equation}

\begin{figure}[ht]
  \centering
  \includegraphics[width=0.6\linewidth]{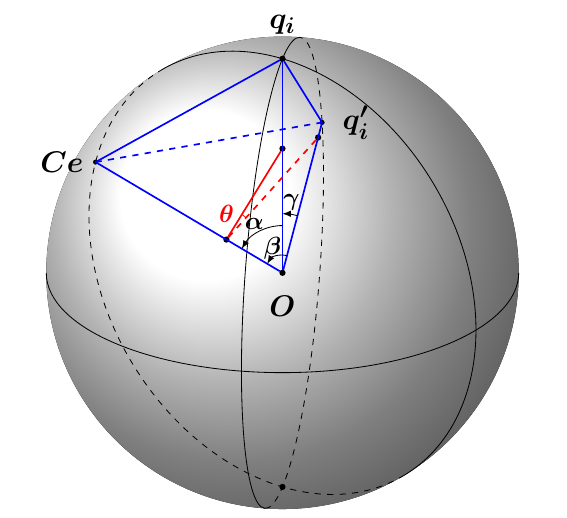}
  \caption{Invariance term schematic. $O$ is the center of the unit hypersphere, $Ce$ is the $\ell_{2}$-normalized one-class center, $q_{i}$ and $q_{i}^{\prime}$ are $\ell_{2}$-normalized projected vectors, $\theta$ is the dihedral angle between plane $CeOq_{i}$ and $CeOq_{i}^{\prime}$, $\alpha$ and $\beta$ are one-class errors, and $\gamma$ is the contrastive error.}
  \label{loss_visual}
\end{figure}

Therefore, the main intuition behind our model is that a positive correlation exists between one-class classification and contrastive learning, allowing their objectives to be achieved simultaneously through a single loss function without the need for hyper-parameter tuning.
Considering ${\rm sim}(u,v) = u^{T}v/\Vert u \Vert_{2}\Vert v \Vert_{2}$ denotes cosine similarity between $u$ and $v$, we define the invariance term $\mathcal{L}_{Inv}$ between $\ell_{2}$-normalized $Q=\left\{q_1,q_2,\dots,q_N\right\}$ and $Q^{\prime}=\left\{q^{\prime}_1,q^{\prime}_2,\dots,q^{\prime}_N\right\}$ as:
\begin{equation}
\begin{aligned}
\mathcal{L}_{Inv}(Q,Q^{\prime}) &= \frac{1}{N}\sum_{i=1}^{N}\left\{\left[1-{\rm sim}(q_{i},Ce)\right] + \left[1-{\rm sim}(q_{i}^{\prime},Ce)\right] \right\} \\
&= \frac{1}{N}\sum_{i=1}^{N}\left\{ 2-{\rm sim}(q_{i},Ce)- {\rm sim}(q_{i}^{\prime},Ce) \right\}, 
\end{aligned}
\label{invariance}
\end{equation}
where $Ce$ is the $\ell_{2}$-normalized one-class center defined by:
\begin{equation}
Ce(Q,Q^{\prime}) = \frac{1}{2N}\sum_{i=1}^{N}(q_{i} + q_{i}^{\prime}).
\label{center}
\end{equation}
Here, $Ce$, $q_{i}$, and $q_{i}^{\prime}$ are distributed on the unit hypersphere after normalization.
According to Eq.~\ref{Deep_SVDD_loss} $\mathcal{L}_{svdd}$, minimizing $\mathcal{L}_{Inv}(Q,Q^{\prime})$ brings $q_{i}$ and $q_{i}^{\prime}$ closer to $Ce$, thereby achieving the one-class classification-based normality assumption.
Meanwhile, on the unit hypersphere, $\mathcal{L}_{Inv}(Q,Q^{\prime})$ and $\mathcal{L}_{sim}$ are related as follows:
\begin{equation}
\mathcal{L}_{Inv}(Q,Q^{\prime}) \geq 1+\mathcal{L}_{sim}(Q,Q^{\prime}),
\label{relation}
\end{equation}
which becomes tighter as $\mathcal{L}_{Inv}(Q,Q^{\prime})$ decreases.
Also, note that minimizing the $\mathcal{L}_{Inv}(Q,Q^{\prime})$ shrinks an upper bound of contrastive errors $\mathcal{L}_{sim}(Q,Q^{\prime})$, fulfilling the contrastive learning-based normality assumption. 
As shown in Fig.~\ref{loss_visual}, on the unit hypersphere, the angle $\alpha$/$\beta$/$\gamma$ is proportional to the Euclidean distance $l_{q_{i}Ce}$/$l_{q_{i}^{\prime}Ce}$/$l_{q_{i}q_{i}^{\prime}}$ between two points.
According to the triangle inequality, the relationship between the three Euclidean distances is $l_{q_{i}Ce} + l_{q_{i}^{\prime}Ce}\geq l_{q_{i}q_{i}^{\prime}}$. 

Therefore, we are maximizing the cosine similarity between $q_{i}$, $q_{i}^{\prime}$, and $Ce$ in Eq.~\ref{invariance}, which is an upper bound on the sequence contrastive learning errors between $q_{i}$ and $q_{i}^{\prime}$.
$\alpha$, $\beta$ and $\gamma$ are related as follows:
\begin{equation}
{\rm cos}\gamma = {\rm cos}\alpha{\rm cos}\beta + {\rm sin}\alpha{\rm sin}\beta{\rm cos}\theta,
\label{angle}
\end{equation}
where $\theta$ is the dihedral angle. 
According to Eq.~\ref{angle},
when $\alpha \to 0$ and $\beta \to 0$, ${\rm cos}\gamma \to 1$.
Therefore, Eq.~\ref{relation} becomes tighter as $\mathcal{L}_{Inv}(Q,Q^{\prime})$ becomes smaller, which was also verified in our experiments.
On the unit hypersphere, the formal proof is as follows:

\begin{equation}
\begin{aligned}
\mathcal{L}_{Inv}(q_{i},q_{i}^{\prime})&=\left[1-{\rm sim}(q_{i},Ce)\right] + \left[1-{\rm sim}(q_{i}^{\prime},Ce)\right] \\
&\propto \alpha + \beta \\
&\propto l_{q_{i}Ce} + l_{q_{i}^{\prime}Ce} \\
&=\sqrt{\|q_{i}-Ce\|^{2}} +\sqrt{\|q_{i}^{\prime} - Ce\|^{2}} \\
&\geq \sqrt{\|q_{i} - q_{i}^{\prime}\|^{2}} \\
&\propto \gamma \\
&\propto 1-{\rm sim}(q_{i},q_{i}^{\prime}) \\
&=1+\mathcal{L}_{sim}(q,q^{\prime}),
\end{aligned}
\label{prove}
\end{equation}
here, $l_{*}$ are the Euclidean distances. $\mathcal{L}_{sim}$ is the contrastive error that metrics the agreement between positive pairs. Similarly, there is a relationship expressed by Eq.\ref{relation} in set Q.

In this paper, we treat representations $q_{i}$ and reconstructed representations $q_{i}^{\prime}$ as positive pairs to learn shared information between different time steps of time series, discarding low-level information that is computationally expensive and unnecessary.
Along with CPC \cite{oord2018representation}, SimCLR \cite{chen2020simple}, and wav2vec \cite{baevski2020wav2vec}, though different in the types of positive pairs, RoCA is essentially computing loss in the representation space.
Therefore, maximizing the cosine similarity of $q_{i}$ and $q_{i}^{\prime}$ in RoCA is a type of negative-sample-free contrastive learning, and we name it ``sequence contrast''. 

\subsubsection{Outlier exposure term}


The invariance term regards the representation $q_i$ projected to the latent space and that $q^{\prime}_{i}$ reconstructed by a Seq2Seq model as positive pairs, and raises a loss function to maximize the cosine similarity between the one-class center $Ce$, $q_i$, and $q^{\prime}_{i}$. It ``squeezes" the border close to the normal samples, as shown in Fig.~\ref{architecture}. 

In contrast to the invariance term, the OE term $\mathcal{L}_{OE}$ characterizes the anomalies mixed in the training set, pushing them away from the normality, increasing the distance between the red and green dotted lines in Fig.~\ref{architecture}, and thus clarifying the boundary. 
It is designed to express the features opposite to the so-called ``normality". 
For example, if a model maximizes the normalized probability $p_i$ and the loss function for normal samples is $\mathcal{L}_{Inv}(X_{i}) = -log(p_i)$, we flip the objective for abnormal ones, resulting in $\mathcal{L}_{OE}(X_{i}) = -log (1-p_i)$.
We maximize the cosine similarity to construct the outlier exposure term $\mathcal{L}_{OE}$, resulting in: 
\begin{equation}
\mathcal{L}_{Inv}(q_{i},q_{i}^{\prime})=2-{\rm sim}(q_{i},Ce)-{\rm sim}(q^{\prime}_{i},Ce),
\label{COCAlossn}
\end{equation}
\begin{equation}
\begin{aligned}
    \mathcal{L}_{OE}(q_{i},q_{i}^{\prime}) &=4-\mathcal{L}_{Inv}(q_{i},q_{i}^{\prime}) \\
    &=2+{\rm sim}(q_{i},Ce)+{\rm sim}(q^{\prime}_{i},Ce),
\end{aligned}
\label{COCAlossa}
\end{equation}
where $0< \mathcal{L}_{Inv}(q_{i},q_{i}^{\prime})\leq 4$. $\mathcal{L}_{OE}$ optimizes for the opposite goal of $\mathcal{L}_{OE}$, that is, to make $q$ and $q^{\prime}$ far enough from $Ce$. It also falls into the range of (0,4] to keep the same magnitude as $\mathcal{L}_{Inv}$.

Assuming that all the labels $y_i$ were known, the joint loss function of normal and abnormal representations should be,
\begin{equation}
\mathcal{L}_{joint}(Q,Q^{\prime}) = \frac{1}{N}\sum_{i=1}^{N} \mu \cdot y_i \cdot \mathcal{L}_{OE}({q_{i},q_{i}^{\prime}})+(1-y_i) \cdot \mathcal{L}_{Inv}(q_{i},q_{i}^{\prime}),
\label{loss}
\end{equation}
where we use a hyperparameter $\mu$ controlling the weight of the sample being anomalous to adapt RoCA to datasets with any proportion of anomalies. When $\mu = 0$, the dataset is clean and therefore $\mathcal{L}_{joint}$ equals $\mathcal{L}_{Inv}$. 

In reality, the labels are unobserved because we suppose all the datasets are corrupted, thus their labels are determined during training with the constraint that
\begin{equation}
\begin{aligned}
\mathcal{Y} = \left\{ y \in \left\{ 0,1 \right\}^{N} : \sum_{i=1}^{N} y_i = \nu N \right\},
\end{aligned}
\label{constraint}
\end{equation}
where $\nu$ refers to the ratio of corrupted data. 
Denoting the parameter set of Eq.~\ref{loss} with $\Theta$, we are facing the minimization issue of 
\begin{equation}
\begin{aligned}
\min _\Theta \min _{y \in \mathcal{Y}} \mathcal{L}_{joint}(\Theta, y)
\end{aligned}
\label{optimization}
\end{equation}
To deal with the constraint discrete optimization problem, we update the $\Theta$ and the labels $\mathcal{Y}$ alternately.
We fix $\mathcal{Y}$ and minimize $\mathcal{L}_{joint}(\Theta)$ to update $\Theta$. Then given $\Theta$, we update $y_i$ under the constraint of Eq.~\ref{constraint}. Therefore, the training anomaly score is defined as
\begin{equation}
\mathcal{S}_{i}^{train} = \mathcal{L}_{Inv}(q_i,q_{i}^{\prime}) - \mathcal{L}_{OE}(q_i,q_{i}^{\prime}),
\label{trainingS}
\end{equation}
which represents the contribution of $y_i$ on minimizing Eq.~\ref{loss}.
We sort these anomaly scores and take the top $\nu$ of the labels $y_i$ to 1, and the remainder to 0. In the first several epochs, only $\mathcal{L}_{Inv}$ is applied to avoid initial bias. Assuming all involved losses are bounded from below, block coordinate descent converges to a local optimum as each update improves the loss. The training process is shown in Algorithm~\ref{plusal}.

\renewcommand{\algorithmicrequire}{\textbf{Input:}}  
\renewcommand{\algorithmicensure}{\textbf{Output:}} 
\begin{algorithm}[htb]
  \caption{Training process in contaminate data} 
  \label{plusal}  
  \begin{algorithmic}[0]
    \Require
      Contaminated training set $\mathcal{D}$, hyperparameter $\nu$
    \For{$Epoch$}
        \ForAll{ Mini batch $ \mathcal{M}$}
            \State Calculate the anomaly score ${\mathcal{S}_{i}^{train}}$ for $X_i \in \mathcal{M}$.
            \State Estimate the label $y_i$ according to ${\mathcal{S}_{i}^{train}}$ and $\nu$.
            \State Update the parameters $\Theta$ by minimizing ${\mathcal{L}_{joint}(y)}$.
        \EndFor
    \EndFor
  \end{algorithmic}
\end{algorithm}

\subsubsection{Variance term}
We note that contrastive learning has two key properties: alignment and uniformity \cite{wang2020understanding} (detail in Section~\ref{related}).
The latter keep the hypersphere from collapse with the help of negative pairs, i.e., avoiding the collapsed solution that outputs a constant.
Nevertheless, uniformity somewhat contradicts the aim of one-class classification \cite{sohn2020learning}, because the latter is to bring representations closer to the center on the unit hypersphere, while some representations may be instead pulled far away by uniformity. 
In the variance term, RoCA removes negative pairs to avoid performance degradation caused by pushing away negative pairs that are both normal. Therefore, a mechanism to maintain uniformity instead of the negative pairs is urgently needed.

Inspired by \cite{bardes2022vicreg,chong2020simple}, RoCA defines the variance term $\mathcal{L}_{Var}$ as a hinge function on the standard deviation of the projected vectors $q_{i}$ to keep the uniformity of the representations:
\begin{equation}
\mathcal{L}_{Var}(Q) = \frac{1}{N}\sum_{i=1}^{N}\max\left\{0,\zeta - \sqrt{{\rm Var}(q_{i})+\varepsilon}\right\},
\label{variance}
\end{equation}
where $\zeta$ is a constant target value of the standard deviation, and $\varepsilon$ is a small scalar to prevent instabilities.
In our experiments, $\zeta$ is set to $1$, and $\varepsilon$ is set to $10^{-4}$.
On the other hand, according to the research in Deep SVDD, selecting an appropriate one-class center can mitigate the problem of hypersphere collapse. 
In RoCA, the one-class center $Ce$ is ensured to be non-zero in any dimension and is only updated during the initial epochs, because experiments show that an unfixed $Ce$ would make the network easily converge to a trivial solution.

Finally, the overall loss function of RoCA is the sum of the joint terms and the variance term:
\begin{equation}
\mathcal{L}_{RoCA} = \mathcal{L}_{joint}(Q,Q^{\prime})
+ \frac{\lambda}{2}(\mathcal{L}_{Var}(Q) + \mathcal{L}_{Var}(Q^{\prime})),
\label{coca_loss}
\end{equation}
where $\lambda$ is the hyper-parameter controlling the contribution of the variance term.
Besides, we proposed its preliminary version in our previous work, namely COCA without introducing OE terms, whose loss function is 
\begin{equation}
\mathcal{L}_{COCA} = \mathcal{L}_{Inv}(Q,Q^{\prime})
+ \frac{\lambda}{2}(\mathcal{L}_{Var}(Q) + \mathcal{L}_{Var}(Q^{\prime})),
\label{previous_loss}
\end{equation}
where $\lambda$ is the hyper-parameter controlling the contribution of the variance term.
The overall algorithm is summarized in Algorithm~\ref{pseudo}.
\begin{algorithm}[htb]
  \caption{RoCA algorithm.} 
  \label{pseudo}  
  \begin{algorithmic}[0]

    \Require
       A set of augmented time series $\left\{{\bf X}_{i}\right\}_{i=1}^{N}$, batch size $N$, structure of $f,g,h,p$,
      constant $\nu, \mu, \zeta, \varepsilon, \lambda$.
    \Ensure
       Parameters of the network $f,g,h$, and $p$.
       \For{sampled batch $\left\{{\bf X}_{i}\right\}_{i=1}^{N}$ in $\mathcal{D}$}
            \ForAll{$i \in [1,N]$}
                   \State ${\bf Z}_{i}=f({\bf X}_{i})$
                   \State ${\bf Z}_{i}^{\prime}=h(g({\bf Z}_{i}))$
                   \State $q_{i}=p({\bf Z}_{i})$
                   \State $q_{i}^{\prime}=p({\bf Z}_{i}^{\prime})$

           \State $Ce = \frac{1}{2N}\sum_{i=1}^{N}(q_{i} + q^{\prime}_{i})$
           \State \textbf{define} ${\rm sim}(u,v)$ \textbf{as} ${\rm sim}(u,v)=u^{T}v/\Vert u \Vert_{2}\Vert v \Vert_{2}$
               \State $\mathcal{L}_{Inv}(q_i,q_i^{\prime})$ = $ 2-{\rm sim}(q_{i},Ce)-{\rm sim}(q^{\prime}_{i},Ce)$
               \State $\mathcal{L}_{OE}(q_i,q_i^{\prime})$ = $ 2+{\rm sim}(q_{i},Ce)+{\rm sim}(q^{\prime}_{i},Ce)$
               \State $\mathcal{S}_{i}^{train}(q_i,q_i^{\prime})$ = $\mathcal{L}_{Inv}(q_i,q_i^{\prime}) - \mathcal{L}_{OE}(q_i,q_i^{\prime})$
            \EndFor
            \State Sort Q according to $\mathcal{S}_{i}^{train}(q_i,q_i^{\prime})$
            \ForAll{$i \in [1, \nu \cdot N]$}
                \State $y_i$ = 1 
            \EndFor
            \ForAll{$i \in [\nu \cdot N +1,\nu\cdot N]$}
                \State $y_i$ = 0 
            \EndFor
            \State 
                $\mathcal{L}_{joint}(Q,Q^{\prime}) = \frac{1}{N}\sum_{i=1}^{N} \mu \cdot y_i \cdot \mathcal{L}_{OE}({q_{i},q_{i}^{\prime}})$
            \Statex \qquad \qquad \qquad \quad $+(1-y_i) \cdot \mathcal{L}_{Inv}(q_{i},q_{i}^{\prime})$
            \State $\mathcal{L}_{Var}(Q) = \frac{1}{N}\sum_{i=1}^{N}\max\left\{0,\zeta - \sqrt{{\rm Var}(q_{i})+\varepsilon}\right\}$
            \State $\mathcal{L}_{Var}(Q^{\prime}) = \frac{1}{N}\sum_{i=1}^{N}\max\left\{0,\zeta - \sqrt{{\rm Var}(q_{i}^{\prime})+\varepsilon}\right\}$
            \State $\mathcal{L}_{RoCA} = \mathcal{L}_{joint}(Q,Q^{\prime}) + \frac{\lambda}{2}(\mathcal{L}_{Var}(Q) + \mathcal{L}_{Var}(Q^{\prime}))$
            \State Update networks $f,g,h$, and $p$ to minimize $\mathcal{L}_{RoCA}$ 
        \EndFor
    \State \textbf{return} Parameters of the network $f,g,h$, and $p$
  \end{algorithmic}
\end{algorithm}

\subsection{Anomaly inference}
During training, we use a joint loss function that combines the optimization objective of both normal samples and latent abnormal representations. In practice, anomalies in the testing phase are not necessarily the same as what we have encountered while training. Therefore, we only use the invariance term $\mathcal{L}_{Inv}$ to score anomalies.
\begin{equation}
\mathcal{S}_{i}^{test} = \mathcal{L}_{Inv}(q_i,q_{i}^{\prime}),
\label{testS}
\end{equation}
where $q_i$ and $q_{i}^{\prime}$ are the original and reconstructed representations of each sample ${\bf X}_i$.
Then, the following formula is applied to determine whether ${\bf X}_i$ can be classified as an anomaly:
\begin{equation}
\begin{aligned}  
x_{i}=\left\{
\begin{array}{lcl}
anomaly,& & \mathcal{S}_{i}^{test} > \tau\\
 normal,& & \mathcal{S}_{i}^{test} \leqslant \tau \quad,
\end{array} \right.
\end{aligned}
\label{COCA_detection}
\end{equation}
where $\tau$ is a predefined threshold.

\section{Experiments}\label{Sec4}
This section presents the experimental setup, baselines, RoCA variants, main results, and hyperparameter analysis. The code for reproducing the results of this paper is available at \url{https://github.com/ruiking04/RoCA}.
\subsection{Experimental Setup}
\subsubsection{Datasets} 
RoCA assesses the proposed method using AIOps, UCR, SWaT, and WADI datasets. Some popular datasets like Numenta \cite{lavin2015evaluating}, Yahoo \cite{yahoo}, NASA (SMAP and MSL) \cite{hundman2018detecting}, and SMD \cite{su2019robust} are not considered due to their weakness of simple to solve or mislabeling, according to the research of \cite{wu2021current}.
The datasets considered are as follows.
\begin{itemize}
    \item \textbf{AIOps} \cite{KPI} 
    comprises 29 univariate time-series sub-datasets of well-maintained business cloud key performance indicators from Internet companies including Tencent, eBay, etc. Some references may identify it as ``KPI".
    \item \textbf{UCR} \cite{wu2021current} contains 250 univariate time-series sub-datasets spanning various domains. Each sub-dataset contains only 1 anomaly segment, thus the training set consists only of normal time series.
    \item \textbf{SWaT} \cite{mathur2016swat} encompasses multivariate data from 51 sensors of the Secure Water Treatment (SWaT) testbed under normal and attacked behavioral modes. It contains more long-term pattern-wise anomalies.
    \item \textbf{WADI} \cite{ahmed2017wadi} is an extension of SWaT with 123 sensors and actuators. The pattern-wise anomalies are shorter than those on SWaT.
\end{itemize}

As demonstrated in Table~\ref{datasets}, each time series is segmented into length-$t$ sequences using a sliding window with a time-step $\delta$. 
The table additionally presents the sequence count in the training, validation, and testing sets and the proportion of abnormal samples.

\begin{table}[htbp]
\caption{Summary of TSAD datasets. }
\renewcommand{\arraystretch}{1.10}
\centering
\scalebox{0.9}{
\begin{threeparttable}
\begin{tabular}{lcccc}
\toprule
  & AIOps & UCR & SWaT & WADI  \\ 
\midrule
 Subsets  & 29  & 250 & 1 & 1\\
 Variables & 1 & 1 & 51 & 127\\
 Domain & Cloud KPIs & Various & Waterworks & Waterworks\\
\midrule
 Length  & 16  & 64 & 32  & 32 \\ 
 Time step & 16 & 16 & 16  & 16  \\
\midrule
 Training & 187741 & 330583 & 29699 & 49034 \\
 Validation & 36493 & 175564 & 5624 & 2160\\
 Testing & 182414  & 877355 & 28118 & 10799\\
 Anomalies & 2.92\%  & 0.53\% & 5.96\% & 1.06\%\\
 Tra-Ano$^{1}$ & 3.24\%  & 0\% & 0\% & 0\%\\
\bottomrule
\end{tabular}
\begin{tablenotes}
    \item[1] Tra-Ano is the proportion of anomalies in the training set.
\end{tablenotes}
\end{threeparttable}
}
\label{datasets}
\end{table}

\subsubsection{Baselines}
We conduct experiments over our RoCA against several baselines, including traditional and deep methods based on assumptions and augmentation. We also use a simple baseline to observe the correctness of various metrics. The detailed description of the baselines is as follows:

\begin{itemize}
\item \textbf{Traditional AD Baselines.} We choose five commonly used traditional AD methods: One-Class SVM (OC-SVM) \cite{scholkopf1999support}, Isolation Forest (IF) \cite{liu2012isolation}, Robust Random Cut Forest (RRCF) \cite{guha2016robust}, Spectral Residual (SR) \cite{ren2019time}, and DAMP \cite{lu2022matrix}. 

\item \textbf{Simple Baseline.} Inspired by \cite{kim2022towards}, we design a simple baseline: the Randomized Anomaly Score (RAS) which generates random anomaly scores during training.

\item \textbf{Single Normality Assumptions-based AD Baselines.} Three single assumption based methods are compared, i.e., LSTM Encoder-decoder (LSTM-ED) \cite{malhotra2016lstm}, Deep one-class (Deep SVDD) \cite{ruff2018deep}, Anomaly Transformer \cite{xu2021anomaly}.

\item \textbf{Multiple Normality Assumptions-based AD Baselines.} Two multiple assumption based methods are chosen, AOC \cite{mou2023deep}, and TCC \cite{sohn2020learning,eldele2021time}.

\item \textbf{Augmentation-based AD Baselines.} Finally, a method based on injecting abnormal samples is set: NCAD \cite{carmona2021neural}. We choose the supervised version of NCAD as the baseline in this paper for its better performance than the unsupervised one.
\end{itemize}

It's worth noting that while Deep SVDD is designed for AD in the image data, their underlying concepts remain significant. 
Therefore, previous work \cite{wang2023deep,mou2023deep,carmona2021neural} has adapted it to the temporal domain by employing Conv1D to implement its autoencoder architecture.
Regarding TCC based on \cite{sohn2020learning}, we conduct representation learning during the pre-training phase using TS-TCC \cite{eldele2021time} and anomaly detection during fine-tuning with the principles of Deep SVDD.

\subsubsection{Metric}

In the time series anomaly detection field, various metrics are used to evaluate the performance of models in detecting point and pattern-wise anomalies. Traditional Point-Wise (PW) metrics tend to underestimate the detection capability of finding pattern-wise anomalies. Therefore, many alternative metrics such as the NAB Score \cite{lavin2015evaluating} and the Point Adjusted (PA) metrics \cite{xu2018unsupervised} are proposed, between which the latter is more popular due to their simpler calculation compared to NAB scores. 

PA metrics assume that if any point within an anomaly window is flagged as anomalous, all points within that window are labeled as true positives. Despite this, other approaches have raised concerns that PA may overestimate anomaly detection (AD) performance. Consequently, alternative metrics such as RPA \cite{hundman2018detecting}, PA\%K \cite{kim2022towards}, and affiliation \cite{huet2022local} metrics have been proposed. During our experiments, it was observed that affiliation metrics also exhibit a tendency to overestimate, with even random anomaly scores yielding high affiliation F1 scores,  which will be reported in Section \ref{main_result}.
PA\%K is a method that falls between PW and PA, where K determines the proportion of the window that must be identified as anomalous to classify the entire window as a true positive. PW and PA correspond to the cases where K equals 0 and 100, respectively, in PA\%K. However, selecting an appropriate K is highly complex in terms of time, and the K chosen from training data may not generalize well to the data distribution in testing or real-world applications. On the other hand, RPA consolidates the entire window into a single sample, aligning with our intuition that ``a single point anomaly and a segment anomaly should both count as one anomaly."

In this study, we choose the RPA F1-score as a fair assessment. We also report the experimental results comparing the four metrics PW, PA, affiliation, and RPA of F1-scores. Note that this paper reports on the metrics for the entire dataset, which is a weighted average of F1 scores for each sub-dataset:
\begin{equation}
{\rm F1_{entire}} = \sum_{i=1}^{M}\frac{e_{i}}{E}{\rm F1}_{i},
\label{all_metrics}
\end{equation}
where $M$ is the number of sub-datasets, $E$ is the total number of anomaly segments for the entire dataset, and $e_{i}$ is the number of anomaly segments of the $i$-th sub-dataset.

\subsection{Implementation details}
The 1D-CNN of the TCN encoder has a dropout rate of $0.45$. We adopt a learning rate from $1e-4$ to $5e-4$, weight decay of $5e-4$, $\beta_{1} = 0.9$, and $\beta_{2} = 0.99$ in an Adam optimizer.
Since in UCR, the time series each has only one anomaly segment, we choose the sample with the largest anomaly score as where the anomaly lies. In addition, we perform the early stopping strategy on UCR, as the time series from different domains vary in epochs to convergence. 
For the other datasets, the obtained raw anomaly scores are converted into Z-scores and we search for the optimal anomaly threshold $\tau \in [-3,3]$ according to the $3\sigma$-rule.
To ensure robustness, each method is executed 10 times with distinct random seeds to obtain the mean and standard deviation of the metrics. 
The models are implemented using PyTorch 1.7 and Merlion 1.1.1 \cite{bhatnagar2021merlion}, and trained on an NVIDIA Tesla V100 GPU.

\subsection{Main Results}
\label{main_result}

\begin{table*}[!ht]
\renewcommand{\arraystretch}{1.25}
\centering
\footnotesize
\caption{ Main results$^1$. 
} 

\centering
\setlength{\tabcolsep}{1mm}{
\scalebox{1}{
\begin{threeparttable}
\begin{tabular}{c|ccc|ccc c}
\toprule
\multirow{2}{*}{\diagbox{\textbf{Method}}{\textbf{Metric}}{\textbf{Dataset}}} & \multicolumn{3}{c|}{AIOps} & \multicolumn{3}{c}{UCR} & \multicolumn{1}{c}{\textbf{Univariate Datasets}}\\ \cmidrule{2-7} \cmidrule(lr){8-8}
                        & RPA Pre     & RPA Rec     & RPA F1    & RPA Pre     & RPA Rec     & RPA F1  & Average RPA F1  \\ \cmidrule{1-7} \cmidrule(lr){8-8}
OC-SVM                  & 3.71    & 66.33   & 7.02   & 15.39   & 17.60  & 16.42  & 11.72\\
IF                      & 1.98$\pm$0.00 &69.49$\pm$0.83 & 3.86$\pm$0.07 & 3.47$\pm$0.36   & 6.44$\pm$0.58   & 4.51$\pm$0.44  &  4.19$\pm$0.26 \\
RRCF                    & 1.49$\pm$0.03 &49.13$\pm$0.98 &2.89$\pm$0.06 & 4.29$\pm$0.53 &9.52$\pm$0.99 & 5.91$\pm$0.67 &4.40$\pm$0.37
\\
SR$^2$                      & 4.47    & \textbf{91.02}   & 8.52   & 22.00 & 22.00 & 22.00 & 15.26
\\
DAMP                    & 1.59    & 9.32    & 2.72   & 39.27        & 38.80        & 39.03 & 22.10  \\
RAS                     & 3.12$\pm$0.57 & 23.60$\pm$4.85 & 5.46$\pm$0.83 & 20.88$\pm$3.06 &20.88$\pm$3.06 & 20.88$\pm$3.06 &13.17$\pm$1.95 \\ \cmidrule{1-7} \cmidrule(lr){8-8}
LSTM-ED                 & 7.85$\pm$0.33 & 67.26$\pm$0.94  & 14.05$\pm$0.52 & 24.64$\pm$1.11 & 24.64$\pm$1.11 & 24.64$\pm$1.11 & 19.35$\pm$0.82\\
Deep SVDD               & 23.02$\pm$8.53 & 32.83$\pm$8.08 & 25.53$\pm$7.38 & 48.05$\pm$2.91 & 48.64$\pm$2.59 & 48.34$\pm$2.74  & 36.94$\pm$5.06 \\
AOC                     & 33.28$\pm$3.73 & 55.27$\pm$3.42 & 41.40$\pm$3.28 & 20.99$\pm$0.78 & 21.80$\pm$0.84 & 21.39$\pm$0.80 & 31.39$\pm$2.04 \\
TCC                     &1.67$\pm$0.42  &18.57 $\pm$3.09 & 3.03$\pm$0.66 & 18.15$\pm$5.26 & 1.36$\pm$0.41 & 2.53$\pm$0.75 & 2.78$\pm$0.71 \\
AnoTrans                & 0.19$\pm$0.06 & 22.39$\pm$1.59 & 0.38$\pm$0.13 & 5.99$\pm$1.27 & 6.04$\pm$1.25 & 6.01$\pm$1.26 & 3.20$\pm$0.70
 \\ \cmidrule{1-7} \cmidrule(lr){8-8}
NCAD                    & 28.02$\pm$2.82 &\underline{77.20$\pm$3.14}  & 41.06$\pm$3.32 & 22.24$\pm$2.99 & 22.24$\pm$2.99 & 22.24$\pm$2.99 & 31.65$\pm$3.16 \\ \cmidrule{1-7} \cmidrule(lr){8-8}
COCA                    & \underline{39.95$\pm$8.42} &  49.40$\pm$7.27   &  \underline{43.74$\pm$6.90}& \textbf{56.33$\pm$1.79} & \textbf{56.24$\pm$1.71} & \textbf{56.29$\pm$1.74} & \underline{50.02$\pm$4.32} \\ 
\rowcolor{black!10} RoCA  & \textbf{46.32$\pm$7.16} & 55.45$\pm$4.94  &  \textbf{50.14$\pm$5.08} & \underline{55.64$\pm$2.59} & \underline{55.64$\pm$2.59} & \underline{55.64$\pm$2.59}  & \textbf{52.18$\pm$2.59} \\ 
\midrule
\midrule
\multirow{2}{*}{\diagbox{\textbf{Method}}{\textbf{Metric}}{\textbf{Dataset}}} &  \multicolumn{3}{c|}{SWaT} & \multicolumn{3}{c}{WADI} & \textbf{Multivariate Datasets}\\ \cmidrule{2-7} \cmidrule(lr){8-8}
                        & RPA Pre     & RPA Rec     & RPA F1    & RPA Pre     & RPA Rec     & RPA F1   & Average RPA F1  \\ \cmidrule{1-7} \cmidrule(lr){8-8}
OC-SVM                  & 0.01    & \textbf{97.14}   & 0.02  & 0.02    & \textbf{100.00}   & 0.03   &  0.025  \\
IF                      &  \underline{57.22$\pm$38.46} &8.00$\pm$1.14 & 12.79$\pm$1.94 &  0.44$\pm$0.08 &  42,86$\pm$9.04 & 0.87$\pm$0.16 & 6.83$\pm$1.05\\
RRCF                    & 0.50$\pm$0.03 & \underline{94.86$\pm$3.79} & 0.99$\pm$0.05 & 0.48$\pm$0.04&\underline{91.43$\pm$6.23} & 0.95$\pm$0.09 & 0.97$\pm$0.07\\
DAMP                    &  0.40       & 2.86        & 0.70 & 0.09        & 7.14        & 0.18  & 0.42\\
RAS                     & 7.58$\pm$2.73 &22.29$\pm$10.67 & 10.30$\pm$2.00 & 6.14$\pm$2.84 & 32.86$\pm$12.04 & \underline{9.77$\pm$3.55} & 8.22$\pm$2.42\\ \cmidrule{1-7} \cmidrule(lr){8-8}
LSTM-ED                 &  3.39$\pm$0.03 & 14.29$\pm$0.00 & 5.48$\pm$0.04 & 2.24$\pm$0.35 & 14.29$\pm$0.00 & 3.86$\pm$0.52 &4.67$\pm$0.28\\
Deep SVDD               & 8.62$\pm$12.08 & 30.57$\pm$8.09 & 8.88$\pm$6.97 & 3.54$\pm$1.41 & 32.14$\pm$14.37 & 6.14$\pm$2.09  &7.51$\pm$4.53\\
AOC                     &  34.78$\pm$0.01 & 22.86$\pm$0.01 & \underline{27.59$\pm$0.02} & 0.18$\pm$0.01 & 57.14$\pm$0.02  & 0.36$\pm$0.01 & 13.98$\pm$0.02\\
TCC                     & 1.17$\pm$0.44 & 17.14$\pm$14.85 & 2.06$\pm$0.66 &  \underline{12.40$\pm$29.26} & 45.71$\pm$39.02        & 5.02$\pm$3.79  & 3.54$\pm$2.23\\
AnoTrans                & \textbf{67.26$\pm$10.75} & 22.00$\pm$1.31 & \textbf{32.91$\pm$1.19} &  0.85$\pm$0.01 & 57.14$\pm$0.00  & 1.68$\pm$0.02 & \underline{17.30$\pm$0.61} \\ \cmidrule{1-7} \cmidrule(lr){8-8}
NCAD                    & 4.28$\pm$1.47 & 34.29$\pm$8.38 & 7.54$\pm$2.48 & 4.28$\pm$1.77 & 19.29$\pm$4.57 & 6.84$\pm$2.53 & 7.19$\pm$2.51\\ 
\cmidrule{1-7} \cmidrule(lr){8-8}
COCA                    & 16.05$\pm$13.40 & 29.52$\pm$7.13  & 15.83$\pm$6.81 & 2.47$\pm$1.31 & 37.50$\pm$32.09 & 4.27$\pm$2.04 & 10.05$\pm$4.43 \\ 
\rowcolor{black!10} RoCA  & 23.16$\pm$9.52 & 22.29$\pm$6.23 & 21.05$\pm$5.11 &  \textbf{18.75$\pm$28.92} & 17.14$\pm$7.28 & \textbf{14.45$\pm$6.93}  & \textbf{17.75$\pm$6.02} \\ 
\bottomrule
\end{tabular}
\begin{tablenotes}
    \item[1] Average RPA Presion rate (\%), Recall rate (\%), and F1-score (\%) with standard deviation for baselines and our method on AIOps, UCR, SWaT, and WADI datasets over 10 runs. The best results are in bold, and the suboptimal ones are underlined. 
    \item[2] SR could not be applied in multivariate time series anomaly detection
\end{tablenotes}
\end{threeparttable}
}}
\label{result}
\end{table*}

We report RPA F1-scores in Table~\ref{result}, and there are two vital observations among the results of the four datasets. Firstly, the performance of most methods on the SWaT and WADI datasets indicate significant barriers between anomaly detection in multivariate and univariate time series. Even AnoTrans, specifically designed to deal with the issue of multivariate time series anomaly detection, only achieves a 32.91\% F1-score on SWaT, and the difficulty grows with the dimension of datasets, as the WADI dataset shows. Furthermore, AnoTrans performs poorly on univariate time series datasets, showing differences in the key focuses of multivariate and univariate detection methods.
Secondly, methods exhibit varying sensitivities to different datasets, such as OC-SVM, DAMP, Deep SVDD, and AOC. Between the two univariate datasets, AIOps contains more point-wise anomalies, while each sub-dataset of UCR contains only 1 anomaly, most of which are pattern-wise. On the other hand, AIOps contains a portion of anomalies in the training set, which are taken as normal during the unsupervised learning process. 
Our proposed RoCA achieves satisfying and balanced performance on the two datasets.

By analyzing the performance of different approaches, several conclusions emerge.
Firstly, among the traditional machine learning methods, DAMP and SR get RPA F1 of 39.03\% and 22\% on UCR, surpassing some deep approaches, indicating the shallow approaches could also work well in specific cases such as detecting univariate pattern-wise anomalies. Nevertheless, they are defective in dealing with high-dimensional multivariate time series.
Secondly, AOC and our COCA of the previous version work better than other baselines based on normality assumptions for TSAD, especially on AIOps and SWaT. This implies that techniques incorporating multiple normality assumptions are more aligned with the nature of normal samples. Additionally, the unsatisfactory results of TCC support the argument that pre-training constraints can hinder the performance of two-staged methods.

Lastly, notable performance enhancements are observed with our COCA and RoCA models across AIOps, UCR, and WADI datasets. RoCA outperforms other baselines by at least 5.9\% and 4.8\% in terms of F1 score, and on the UCR dataset, COCA achieves the highest performance, surpassing others by 7.9\%. COCA outperforms RoCA because of the pure training set of UCR, which is not too large and can be carefully labeled. On the other hand, SWaT and WADI are large and high-dimensional datasets, which may be mixed with unknown noises and can also be observed in our ablation study Subsection~\ref{ablation} and hyperparameter analysis Subsection~\ref{Hyperparameters Analysis}.
These underscore the effectiveness and robustness of ensemble techniques that incorporate multiple normality assumptions.

\begin{figure*}[!htb]
\centering

\subfigure[IOpsCompetition]{
\begin{minipage}[t]{0.33\linewidth}
\centering
\includegraphics[width=\linewidth]{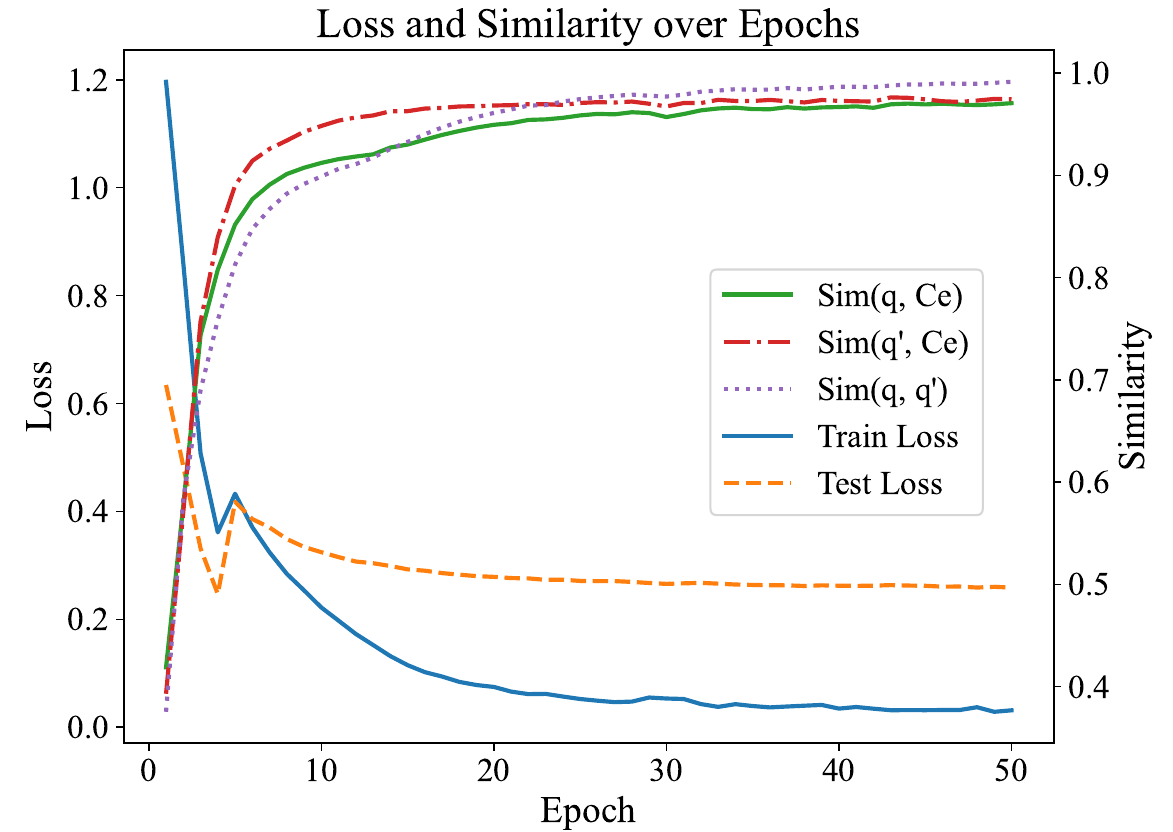}
\label{lckpi}
\end{minipage}
}
\subfigure[UCR]{
\begin{minipage}[t]{0.33\linewidth}
\centering
\includegraphics[width=\linewidth]{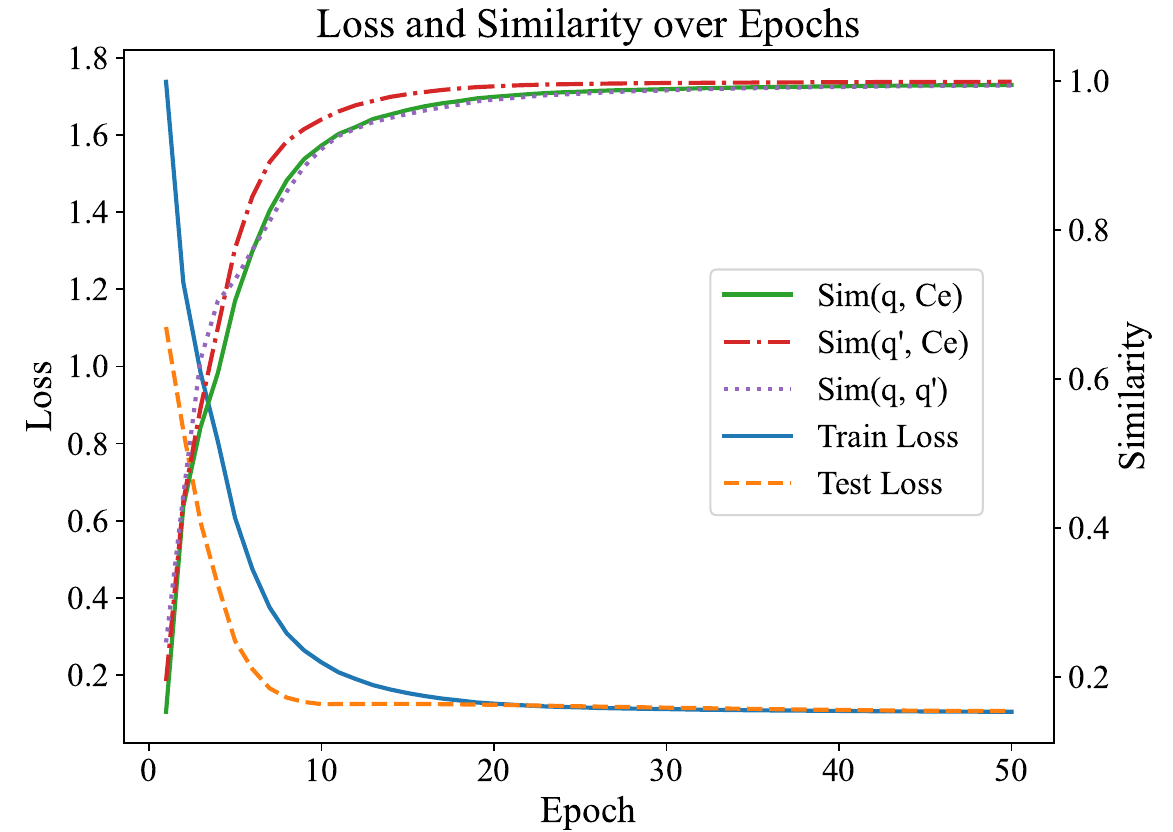}
\label{lcucr}
\end{minipage}
}
\subfigure[SWaT]{
\begin{minipage}[t]{0.33\linewidth}
\centering
\includegraphics[width=\linewidth]{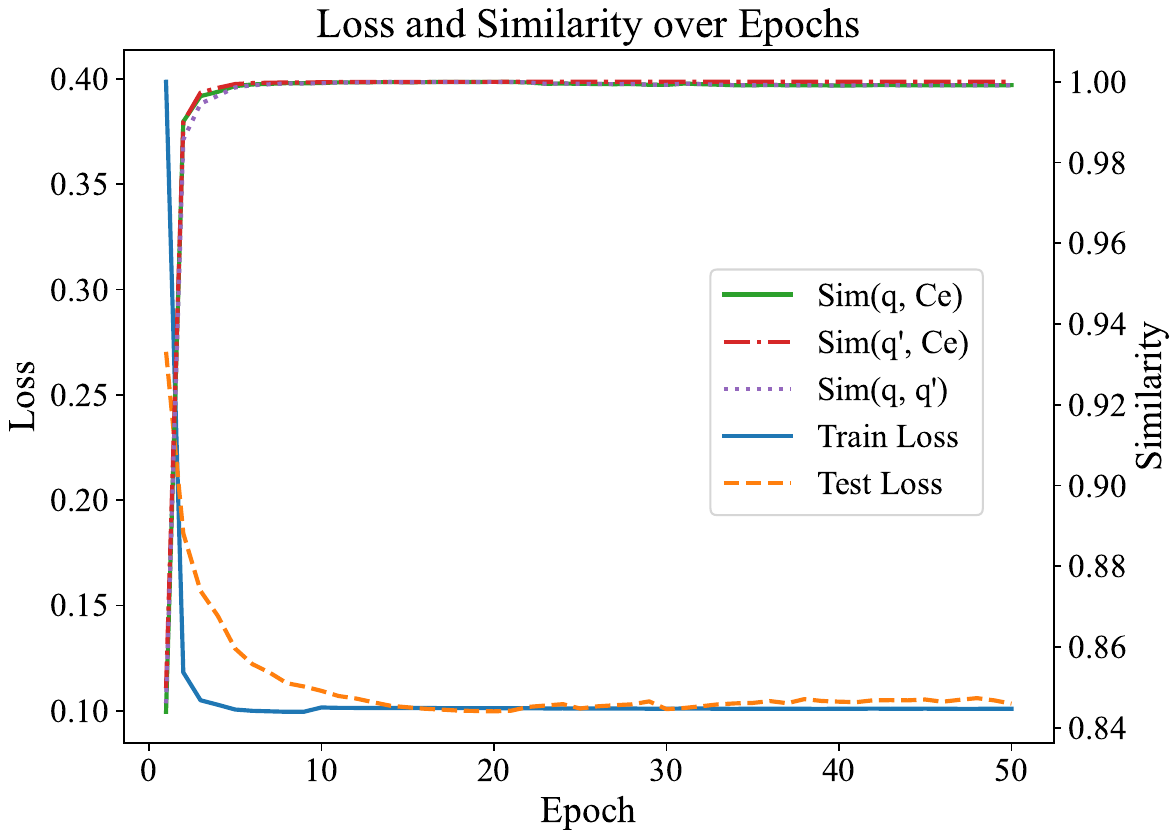}
\label{lcswat}
\end{minipage}
}
\subfigure[WADI]{
\begin{minipage}[t]{0.33\linewidth}
\centering
\includegraphics[width=\linewidth]{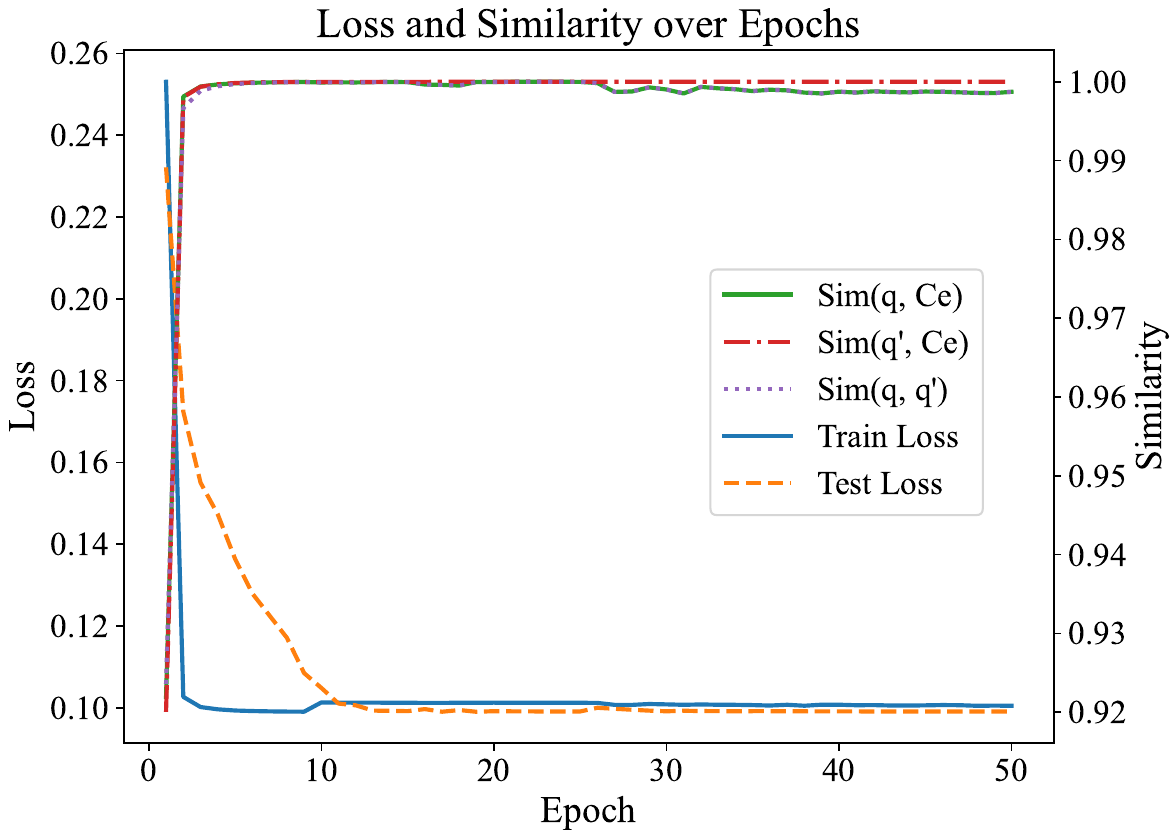}
\label{lcwadi}
\end{minipage}
}

\caption{Loss and the similarity among $q$,$q'$, and the one class center ($c$) during training on the four datasets. The loss function not only brings $q_{i}$ and $q_{i}^{\prime}$ closer to $Ce$, but also reduces the sequence contrastive error ${\rm sim}(q_{i}, q_{i}^{\prime})$.
}
\label{lc_figs}
\end{figure*}

\subsection{Ablation study} \label{ablation}


The RoCA loss function contains three parts, the invariance, OE, and variance terms.
We include the following COCA variants as baselines to demonstrate the effectiveness of individual components in RoCA. Their constructions are shown in Table~\ref{ablation_results}. 
\begin{itemize}
\item \textit{COCA.} The original multiple assumption-based methods consist of the invariance and variance terms, ignoring the corrupted data.

\item \textit{COCAS.} On the basis of COCA, COCAS uses a soft-boundary strategy instead of the OE term to exclude the influence of contaminated data. The soft boundary allows a small portion of samples to fall out of the normal cluster during training. In COCAS, the invariance term is 
\begin{equation}
\mathcal{L}_{InvS}(Q,Q^{\prime}) = Qau + \frac{1}{rN}\sum_{i=1}^{N}\max\left\{0,\mathcal{S}^{train}_{i}-Qau\right\},
\label{soft_boundary_inv}
\end{equation}
where $Qau$ is the $(1-r)$-quantile of the anomaly scores during training, hyper-parameter $r\in(0,1]$ controls the trade-off between $Qau$ and violations of the boundary, i.e. the amount of time series allowed to be mapped outside the boundary.
The loss function of COCAS is
\begin{equation}
\mathcal{L}_{COCAS} = \mathcal{L}_{InvS}(Q,Q^{\prime})
+ \frac{\lambda}{2}(\mathcal{L}_{Var}(Q) + \mathcal{L}_{Var}(Q^{\prime})),
\label{cocaS_loss}
\end{equation}
where $\lambda$ is the hyper-parameter controlling the contribution of the variance term.

\item \textit{RoCA-noV.} It consists of the invariance and OE terms, i.e., removes the variance term from RoCA. 
\end{itemize}


Table~\ref{ablation_results} presents the effectiveness of each component in our proposed RoCA model. Comparing COCA, COCAS, and RoCA, we can observe the effectiveness of different strategies for handling contaminated data. On contaminated AIOps, COCAS uses soft boundaries to exclude the impact of anomalies, but its performance is weaker than that of naive COCA. On the contrary, it works best on SWaT, indicating that the conditions for the effectiveness of the soft-boundary mechanism may be somewhat strict. Meanwhile, RoCA is more effective on AIOps and WADI than the other two, demonstrating the effectiveness of the OE term. COCA performs best on UCR since the UCR dataset is relatively small, which is conducive for publishers to annotate carefully and thus avoids data contamination. On the other hand, on high-dimensional SWaT and WADI, the OE term seems to be effective. We believe that this may be attributed to the difficulty in avoiding noise in the process of collecting large amounts of high-dimensional data.
on the other hand, the comparison between RoCA-noV and RoCA proves the effectiveness of the variance term.

\begin{table*}[!ht]
\centering
\setlength{\tabcolsep}{1mm}{
\caption{RoCA variants and ablation results.}
\label{ablation_results}
\scalebox{1}{
\begin{tabular}{cc|cccc}
\toprule
Variants & Componentss   & AIOps    & UCR           & SWaT & WADI      \\ \midrule
COCA&$\mathcal{L}_{Inv}+\mathcal{L}_{Var}$   &  43.74$\pm$6.90 & \textbf{56.29$\pm$1.74} & 15.83$\pm$6.81 & 4.27$\pm$2.04  \\
COCAS&$\mathcal{L}_{InvS}+\mathcal{L}_{Var}$   & 40.59$\pm$8.11 & 50.94$\pm$2.99 & \textbf{22.73$\pm$8.12} & 3.73$\pm$3.73 \\
RoCA-noV&$\mathcal{L}_{Inv}+\mathcal{L}_{OE}$ &  \underline{49.93$\pm$4.95} & 55.39$\pm$2.19 & 18.86$\pm$7.81 & \underline{6.99$\pm$5.89} \\
RoCA&$\mathcal{L}_{Inv}+\mathcal{L}_{Var}+\mathcal{L}_{OE}$    & \textbf{50.14$\pm$5.08}& \underline{55.64$\pm$2.59} & \underline{21.05$\pm$5.11} & \textbf{14.45$\pm$6.93} \\ \bottomrule
\end{tabular}
}}
\end{table*}

In addition, to verify the validity of the invariance terms in the loss function of RoCA, Fig. \ref{lc_figs} illustrates loss and cosine similarity results for RoCA on the four datasets.
The process of optimizing the loss function $\mathcal{L}$ makes ${\rm sim}(q_{i},Ce) \to 1$, ${\rm sim}(q_{i}^{\prime}, Ce) \to 1$ and ${\rm sim}(q_{i},q_{i}^{\prime}) \to 1$, indicating that the loss function not only makes $q_{i}$ and $q_{i}^{\prime}$ closer to $Ce$, but also minimizes the sequence comparison error ${\rm sim}(q_{i},q_{i}^{\prime})$. This observation is consistent with the derivation of Eq.\ref{prove}.

\begin{figure}[!htb]
\centering

\subfigure[UCR]{
\begin{minipage}[t]{0.9\linewidth}
\centering
\includegraphics[width=\linewidth]{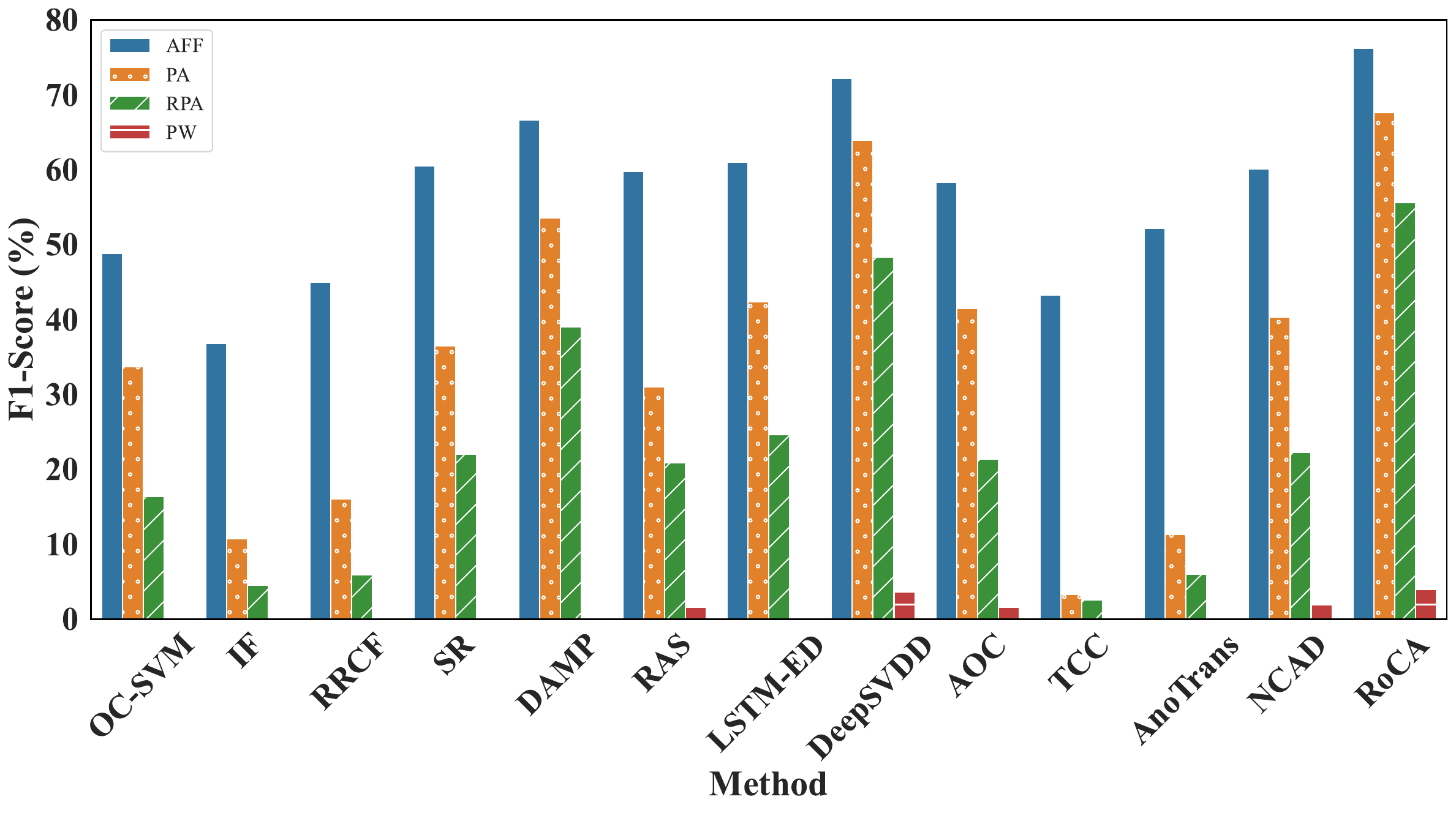}
\label{metricucr}
\end{minipage}
}
\subfigure[SWaT]{
\begin{minipage}[t]{0.9\linewidth}
\centering
\includegraphics[width=\linewidth]{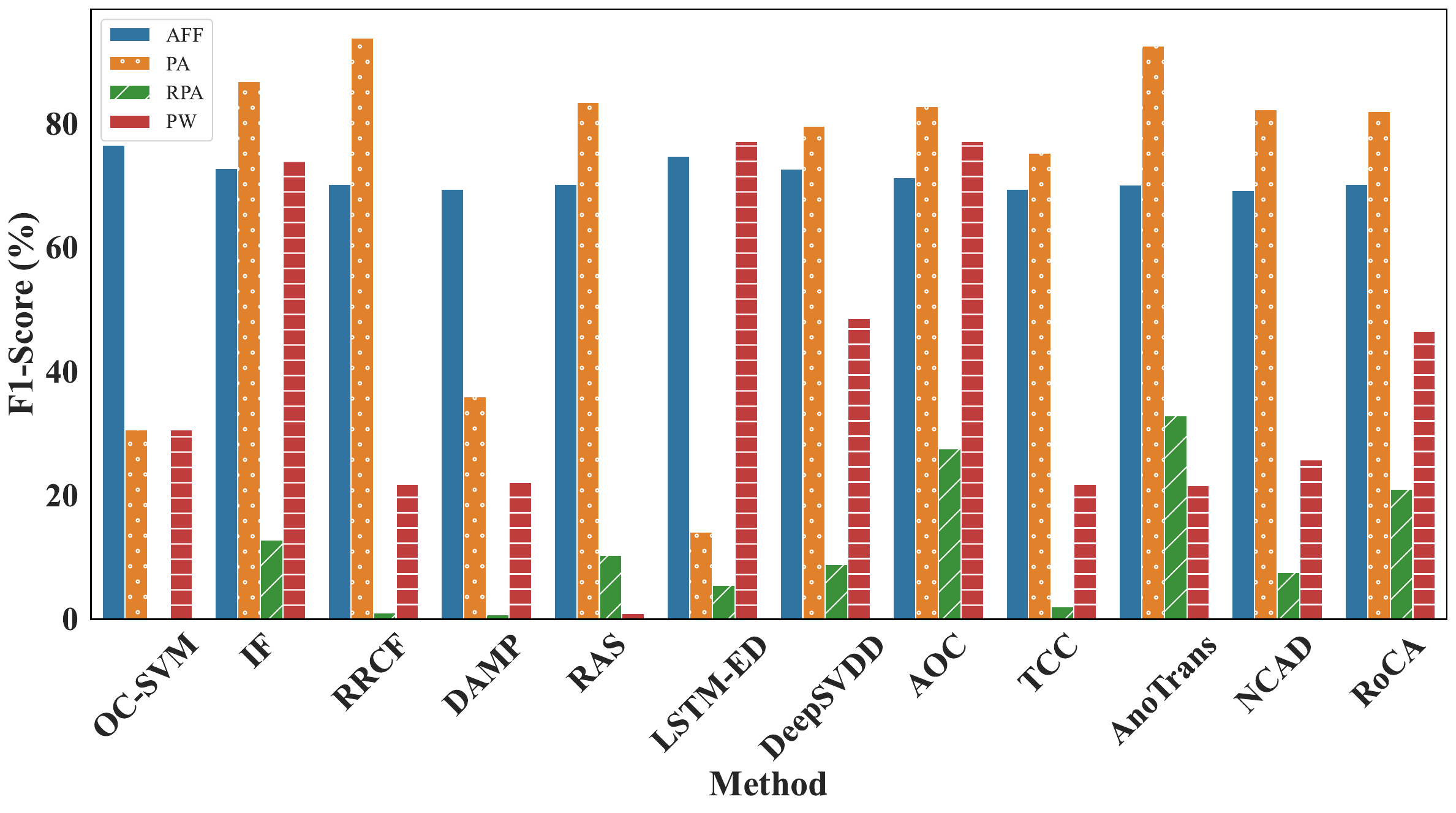}
\label{metricswat}
\end{minipage}
}

\caption{Experimental metric comparison. RAS is a straightforward method of generating random anomaly scores. The left methods are shallow baselines, and the right six are deep ones. (a) is on UCR which includes pattern-wise anomalies. (b) is conducted on multivariate SWaT, where SR could not work. AFF and PA obtain high scores even when testing RAS, while PW is overly harsh in considering that every method performs poorly on UCR. The comparison between the baselines on different metrics and datasets implies that metrics and benchmarks mislead the fair evaluation.}
\label{metric_figs}
\end{figure}

\subsection{Metric Comparison}
We conduct experiments on two datasets with four different metrics, whose results are shown in Fig.~\ref{metric_figs}. 
On the UCR dataset, which contains only one anomaly segment on each sub-dataset, each method performs extremely poorly under the PW metric, with F1 scores reaching single digits. Anomalies on the UCR dataset often appear in the form of segments, while algorithms may not be able to identify every point within them. PW treats each point as a verification object, thus it is too harsh in evaluating all methods.

On both datasets, PA and affiliation metrics report very high F1 scores. Especially the affiliation one, which forms nearly a straight line on the SWaT dataset and could hardly distinguish the performance of various methods. What's worse, our simple baseline RAS even beats many deep models on the thorn SWaT dataset over the two metrics. We attribute this overestimation to the high portion of long-term sequence (pattern) anomalies in this dataset. When evaluating the performance of a method, PA regards each point in an anomalous segment as an anomaly. However, it unreasonably assumes that all of the predictions are correct throughout the whole anomaly segment once it encounters a correctly predicted point. 

Considering that the fairness of metrics is influenced by the types of anomalies, this article cautiously adopts RPA as the metric, and the reported ``F1-scores" are RPA ones.

\subsection{Hyperparameters Analysis}
\label{Hyperparameters Analysis}
\begin{figure}[ht]
\centering

\subfigure[$\mu$ on AIOps]{
\begin{minipage}[t]{0.45\linewidth}
\centering
\includegraphics[width=\linewidth]{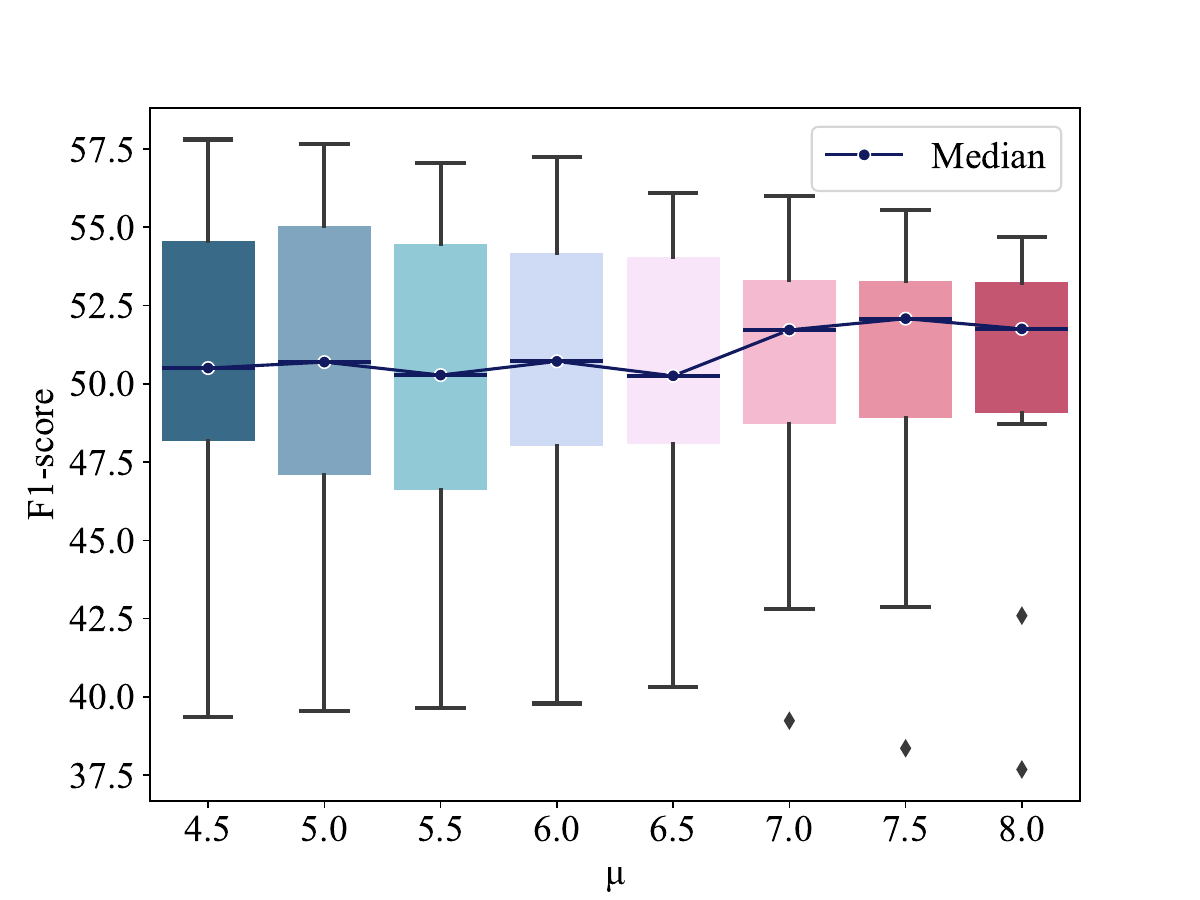}
\label{mu}
\end{minipage}
}
\subfigure[$\nu$ on SWaT]{
\begin{minipage}[t]{0.45\linewidth}
\centering
\includegraphics[width=\linewidth]{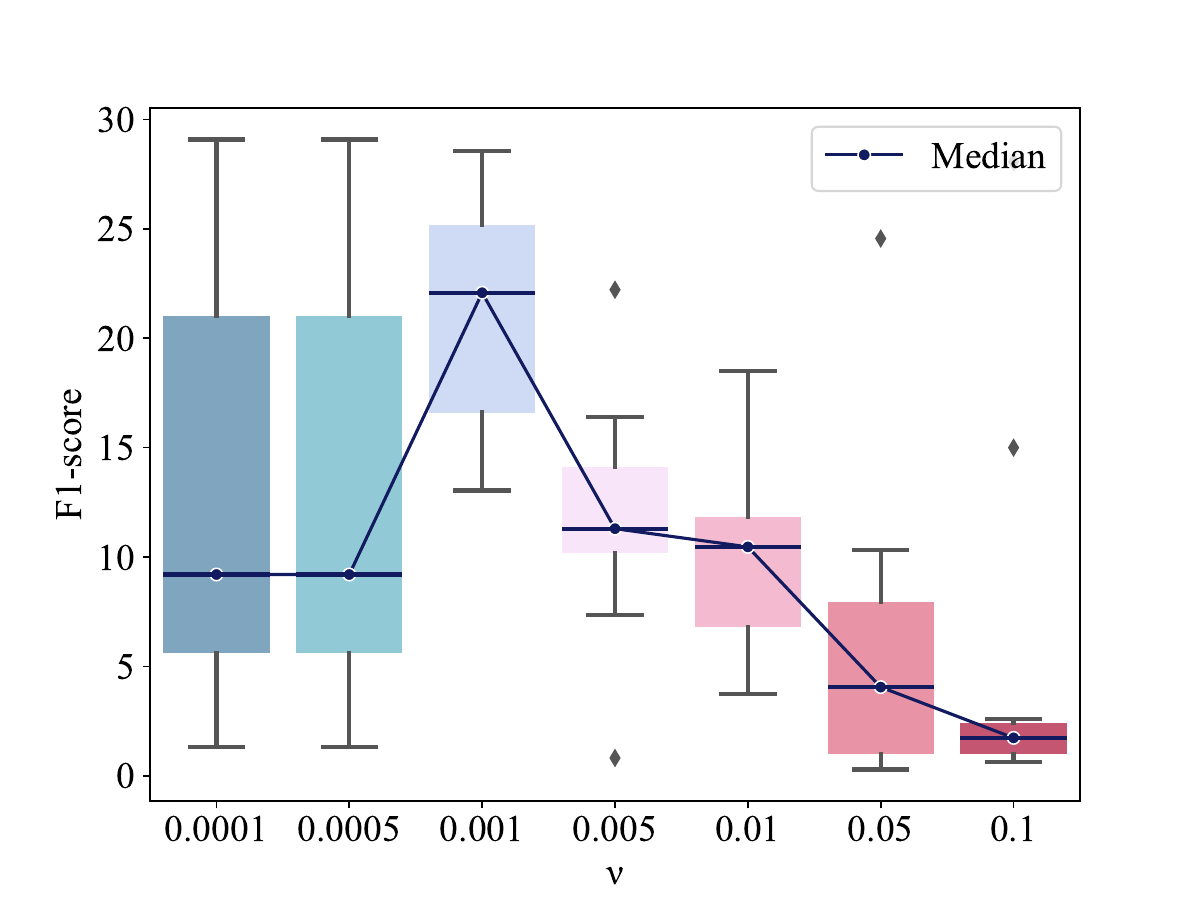}
\label{nu}
\end{minipage}
}

\caption{Sensitivity learning of $\mu$ and $\nu$ on AIOps and SWaT datasets. The box plots present the results of ten times running, including (from the bottom to the top) the minimum, lower quartile, median, upper quartile, and maximum results. The gray diamonds are extreme outliers in the experimental results.}
\label{sensitivity}
\end{figure}

We conducted a sensitivity analysis to study the impact of hyperparameters, including the weight of the OE term ($\mu$) and the proportion of contaminated data ($\nu$),
as shown in Fig.~\ref{sensitivity}.
The y-axis represents the RPA F1 scores of ten times running. 
What's certain is that our model is sensitive to $\mu$ and $\nu$.
Since dataset AIOps contains anomalies in its training set, we study $\mu$ in a range of [4.5,~8.0] on it. 
As shown in Fig.~\ref{mu}, it is evident that the OE term influences the performance in terms of both the average F1 score and its standard deviation. When the weight $\mu$ is set between 4.5 and 6.0, the ten experiments' maximum and minimum values differ even by 10\%. Meanwhile, the medians of the F1 scores are smaller than that when $\mu \geq 7.0$. We regard 7.0 as the best $\mu$, considering the degree of clustering of 50\% of the results, the number of outliers among the ten times running, and the degree that the outliers deviate from the box plot distribution. As $\mu \geq 7.0$ grows, the robustness of the model decreases since larger and more outliers appear at the value of 7.5 and 8.0.

Fig.~\ref{nu} shows the results of varying $\nu$ in a range of 0.0001 to 0.1 on the SWaT dataset. The best performance comes from an assumption that 0.001 proportion of the samples are corrupted. We speculate that this is because data collected in real industrial environments are often mixed with some noise, although this dataset claims it does not contain anomalies. Smaller $\nu$ at 0.0001 and 0.0005 seem the same because very few samples are judged as anomalous and contribute to clarifying the boundaries. When it's set to 0.05 or 0.1, the OE term also improves the performance but seems too strict. However, a much larger $\nu$ may damage the model's functionality and make it overfit to normal data.


\begin{figure*}[ht]
\centering

\subfigure[Visualization on AIOps]{
\begin{minipage}[t]{0.35\linewidth}
\centering
\includegraphics[width=\linewidth]{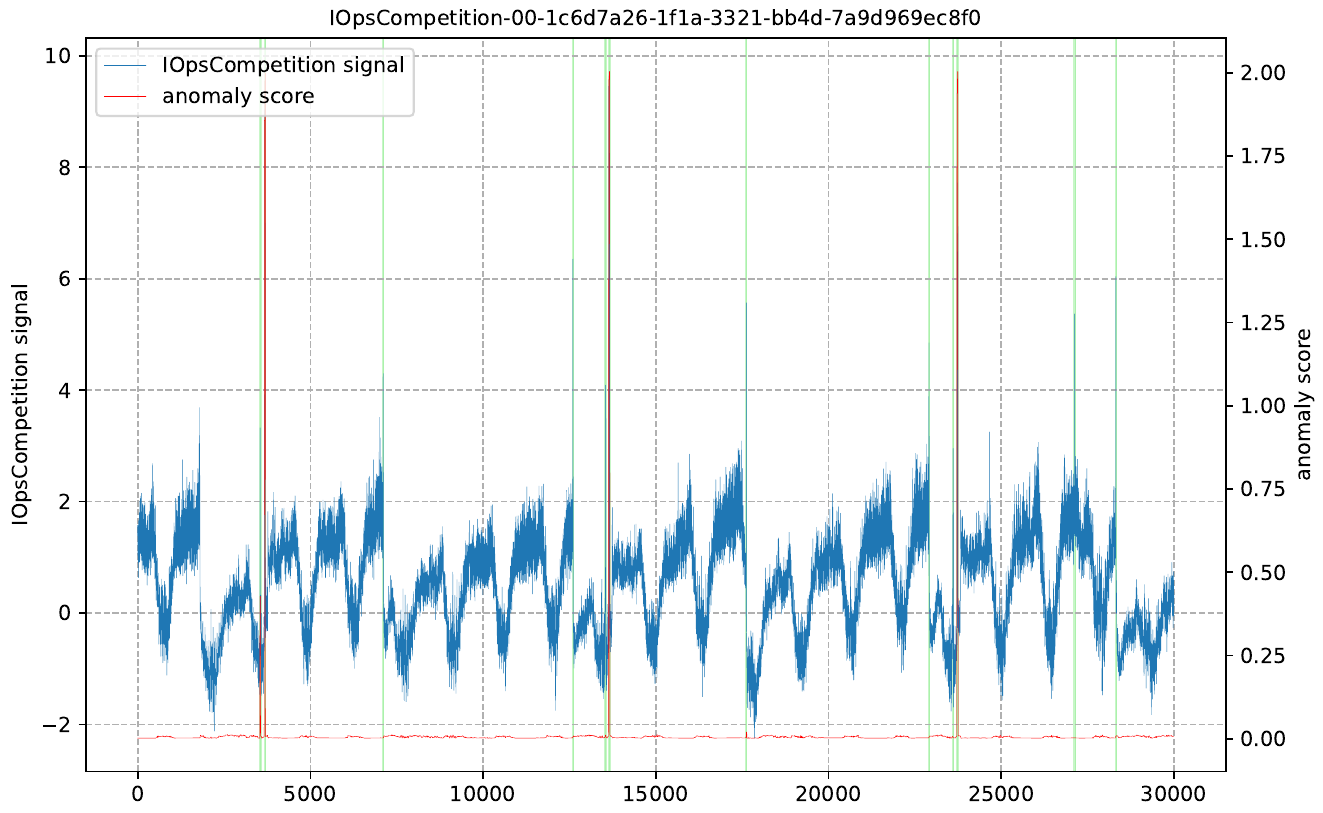}
\label{visualKPI}
\end{minipage}
}
\subfigure[Visualization on UCR]{
\begin{minipage}[t]{0.35\linewidth}
\centering
\includegraphics[width=\linewidth]{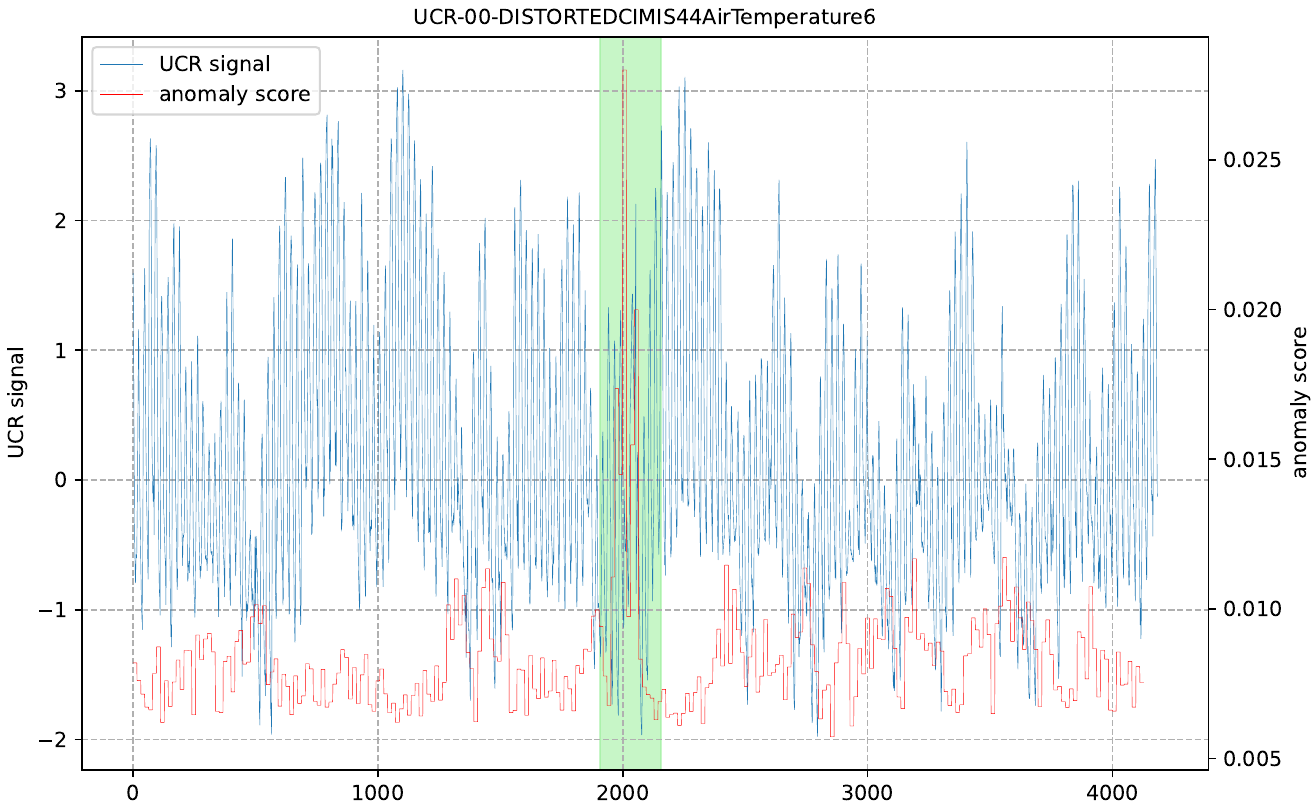}
\label{visualUCR}
\end{minipage}
}
\subfigure[Visualization on SWaT]{
\begin{minipage}[t]{0.32\linewidth}
\centering
\includegraphics[width=\linewidth]{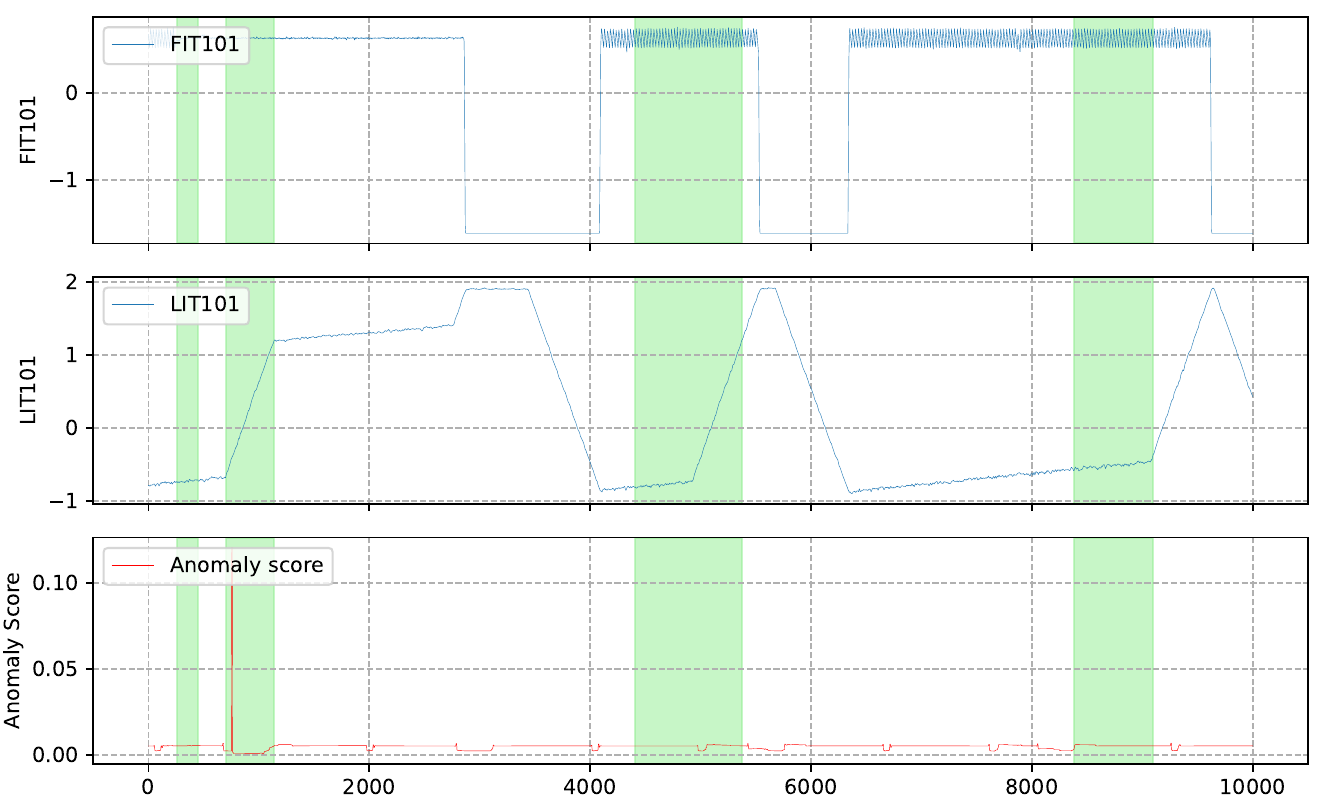}
\label{visualSWaT}
\end{minipage}
}
\subfigure[Visualization on WADI]{
\begin{minipage}[t]{0.34\linewidth}
\centering
\includegraphics[width=\linewidth]{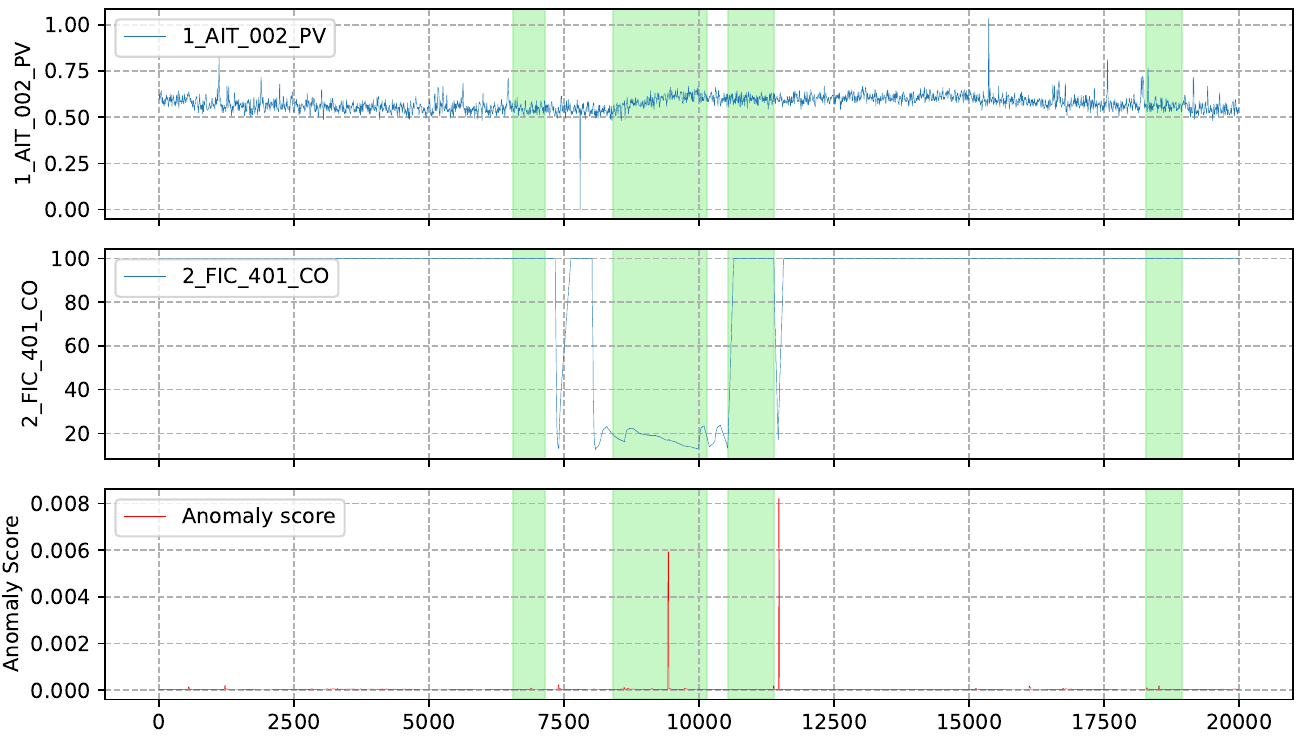}
\label{visualWADI}
\end{minipage}
}

\caption{Visualization on the four datasets. The titles represent the IDs of their subsets, with the x-axis denoting timestamps and the y-axis showing signal values. The blue curves correspond to the original data, while the red curves represent the anomaly scores predicted by RoCA. Ground-truth anomalies are highlighted in green. Each sub-dataset in UCR contains only one anomalous segment. Two dimensions of multivariate datasets are selected for visualization.}
\label{visualization}
\end{figure*}

\subsection{Visualization}
To provide a more intuitive evaluation, we visualize the detection results of our method on the four datasets in Fig.\ref{visualization}. In AIOps, there are numerous point anomalies, and our method can accurately identify the periodic point anomalies within them. In the UCR dataset, each sequence contains only one anomalous segment. As shown, RoCA produces distinctly higher anomaly scores at the corresponding positions, clearly differentiating them from the surrounding points.
For multivariate datasets, we showcase signals from different sensors alongside our predicted anomaly scores, specifically FIT\_101 and LIT\_101 for SWaT, and 1\_AIT\_002\_PV and 2\_FIC\_401\_CO for WADI.
Over a 30,000-point period, the anomaly score remains close to 0 most of the time. RoCA detects two anomalous segments with exceptionally high anomaly scores and also captures the changes in anomaly scores at other anomalous points. 

Overall, our method is capable of addressing both point anomalies and pattern anomalies in most time series. In multivariate time series data, there may be more complex correlation-based anomalies, which require further exploration on our part.

\subsection{Robustness improvement}
In practical applications, the robustness of a model is a key indicator of its ability to maintain stable performance when confronted with data noise, outliers, or complex scenarios.
To comprehensively evaluate the robustness improvement from COCA to RoCA, we introduced more anomalies into the AIOps training set. Specifically, by adjusting the pollution rate $pr$, we added anomalies to the training set in proportion to $pr$ times the original amount. The figures display the robustness experiment results for $pr$ values of 0.5 and 1–7. Since these experiments were not repeated multiple times to obtain average values, they also exhibit a degree of randomness.

\begin{figure}[htb]
  \centering
\subfigure[Precision]{
\begin{minipage}[t]{0.46\linewidth}
\centering
\includegraphics[width=\linewidth]{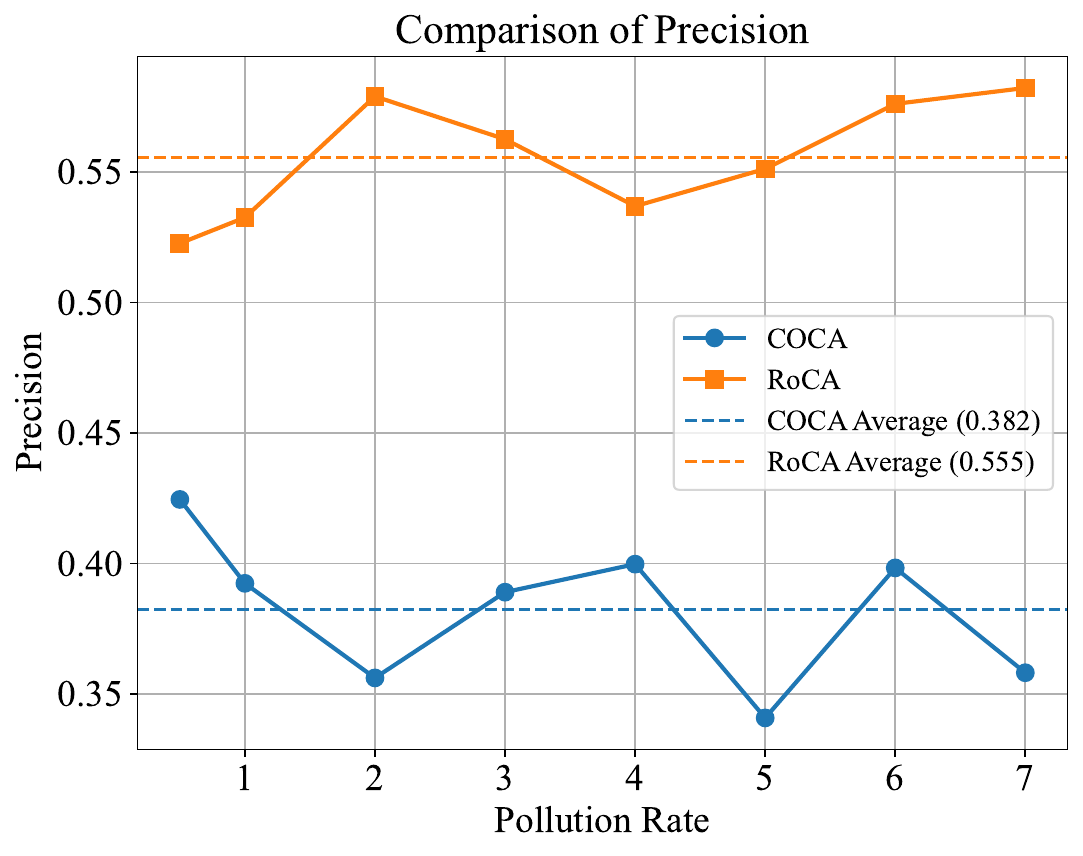}
\label{robust_pre}
\end{minipage}
}
\subfigure[F1-score]{
\begin{minipage}[t]{0.46\linewidth}
\centering
\includegraphics[width=\linewidth]{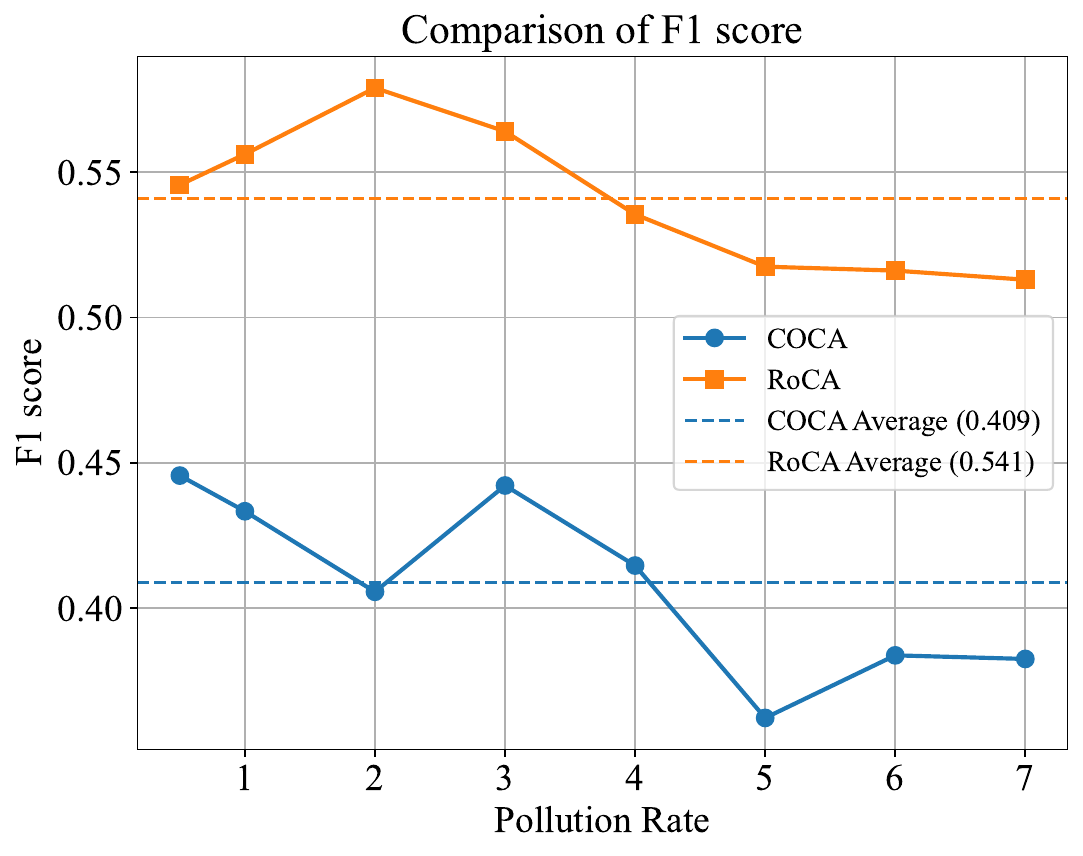}
\label{robust_F1}
\end{minipage}
}
    \caption{Robustness experiment result on AIOps dataset.}
  \label{robust}
\end{figure}
In terms of overall performance, COCA achieves an average precision of 0.38, whereas RoCA significantly enhances this metric to 0.56, indicating that RoCA is more effective at identifying positive samples while reducing misclassification. Furthermore, COCA attains an average F1-score of 0.41, while RoCA improves it to 0.54, demonstrating a better balance between precision and recall.

Regarding performance stability, COCA exhibits considerable fluctuations in precision and F1-score across different pollution rates. For instance, at a pollution rate of 0.5, its precision and F1-score are 0.42 and 0.45, respectively. However, as the pollution rate increases to 7, these metrics drop to 0.36 and 0.38.
In contrast, RoCA demonstrates significantly smaller performance variations. Even at higher pollution rates (e.g., 7), its precision and F1-score remain stable at approximately 0.58 and 0.52. This highlights RoCA’s ability to maintain consistent performance across varying noise levels and its stronger robustness against latent anomalies in the training set, making it well-suited for practical applications on unlabeled or coarsely processed data.

\section{Conclusions}\label{section conclusion}
We propose the robust RoCA, the first attempt at a multi-assumption based time series anomaly detection method for contaminated data, to our best knowledge.
It integrates the normality assumptions of contrastive learning and one-class classification into a bind one, clarifies the essence of contrastive learning, and introduces a new negative-sample-free approach termed ``sequence contrast." 
Furthermore, it introduces a crucial outlier exposure term in the loss function to leverage the latent anomalies mixed in the training set, clarifying the boundary of normal and abnormal representations. Experiments on both uni- and multivariate datasets demonstrate the effectiveness and robustness of RoCA.
We hope that our work can deepen the understanding of time-series anomaly detection, expand the possibilities of integrating various anomaly detection methods, and enhance the robustness of anomaly detection in practical applications.

The normality of multivariate time series has not been roundly characterized according to the experimental results on SWaT and WADI datasets. High-dimensional data mining and the normality of the relationship between multiple dimensions will be explored in our future work. On the other hand, generating possible anomalies based on potentially contaminated data to balance samples is also our future research direction.














\bibliographystyle{IEEEtran}

\bibliography{ref.bib}






\end{document}